\pdfoutput=1

\documentclass[11pt]{article}
\usepackage{amssymb}
\usepackage{listings}

\usepackage[final]{acl}

\usepackage{amsmath,amsfonts,bm}

\def\eqref#1{equation~\ref{#1}}

\def\1{\bm{1}}

\DeclareMathAlphabet{\mathsfit}{\encodingdefault}{\sfdefault}{m}{sl}
\SetMathAlphabet{\mathsfit}{bold}{\encodingdefault}{\sfdefault}{bx}{n}

\makeatletter
\renewcommand{\@fnsymbol}[1]{\ensuremath{%
   \ifcase#1\or 
   \circ\or    
   \dagger\or   
   \ddagger\or          
   \mathsection\or 
   \mathparagraph\or 
   \|\or           
   **\or          
   \dagger\dagger\or 
   \ddagger\ddagger\else 
   \@ctrerr  
   \fi
}}
\makeatother

\usepackage{times}
\usepackage{latexsym}
\usepackage{booktabs}    
\usepackage{multirow}    
\usepackage{adjustbox}
\usepackage{amsmath}
\usepackage{bm}
\usepackage[most]{tcolorbox}
\usepackage{bbm}
\usepackage{enumitem}
\usepackage{graphicx}
\usepackage{tikz}
\usepackage{xcolor}
\usepackage[capitalize]{cleveref}
\usepackage{float}
\usepackage{enumitem}
\usepackage[T1]{fontenc}

\usepackage[utf8]{inputenc}

\usepackage{microtype}

\usepackage{inconsolata}
\usepackage{amsmath,amsthm}

\newcommand{\EE}{\mathcal{E}}
\newcommand{\RR}{\mathcal{R}}

\newcommand{\Gtruth}{\mathcal{G}_{\textsc{t}}}
\newcommand{\Etruth}{\mathcal{E}_{\textsc{t}}}
\newcommand{\Rtruth}{\mathcal{R}_{\textsc{t}}}
\newcommand{\Ftruth}{\mathcal{F}_{\textsc{t}}}
\newcommand{\Gobs}{\mathcal{G}_{\textsc{o}}}
\newcommand{\Eobs}{\mathcal{E}_{\textsc{o}}}
\newcommand{\Robs}{\mathcal{R}_{\textsc{o}}}
\newcommand{\Fobs}{\mathcal{F}_{\textsc{o}}}
\newcommand{\Gtrain}{\mathcal{G}_{\textsc{train}}}
\newcommand{\Etrain}{\mathcal{E}_{\textsc{train}}}
\newcommand{\Rtrain}{\mathcal{R}_{\textsc{train}}}
\newcommand{\Ftrain}{\mathcal{F}_{\textsc{train}}}
\newcommand{\Gtest}{\mathcal{G}_{\textsc{test}}}
\newcommand{\Etest}{\mathcal{E}_{\textsc{test}}}
\newcommand{\Rtest}{\mathcal{R}_{\textsc{test}}}
\newcommand{\Ftest}{\mathcal{F}_{\textsc{test}}}
\newcommand{\txt}{\Sigma^*}
\newcommand{\labl}{{l}}

\usepackage{graphicx}

\def\ultra{{\textsc{Ultra}}}
\def\ours{{\textsc{Semma}}}
\def\mgnn{{\textsc{Motif}}}
\def\trix{{\textsc{Trix}}}
\def\prolink{{\textsc{ProLINK}}}

\def\rgone{{\mathcal{G}_{R}^{\textsc{str}}}}
\def\rgtwo{{\mathcal{G}_{R}^{\textsc{text}}}}
\def\nollm{{\textsc{rel\_name}}}
\def\cleaned{{$\textsc{llm\_rel\_name}$}}
\def\desc{{$\textsc{llm\_rel\_desc}$}}
\def\cs{{$\textsc{combined\_sum}$}}
\def\cavg{{$\textsc{combined\_avg}$}}

\newtheorem{definition}{Definition}[]

\definecolor{kleinblue}{RGB}{0, 47, 167}
\definecolor{airforceblue}{rgb}{0.36, 0.54, 0.66}
\definecolor{finalblue}{RGB}{108, 142, 191}
\definecolor{igreen}{HTML}{228B22}
\usetikzlibrary{shadows}

\DeclareRobustCommand{\circled}[1]{%
  \tikz[baseline=(char.base)]{%
    \node[shape=circle, fill=finalblue, draw=white, thick,  
          inner sep=1pt] (char) {\textcolor{white}{#1}};%
  }
}

\tcbuselibrary{skins, breakable} 

\tcbset{
  promptblueprint/.style={
    enhanced, 
    colback=blue!5!white, 
    colframe=blue!60!black, 
    coltitle=black,
    fonttitle=\bfseries\large,
    boxrule=1pt,
    arc=2pt, 
    left=8pt,
    right=8pt,
    top=8pt, 
    bottom=8pt,
    breakable, 
    pad at break=2mm,
    title after break={\centering --- Model Prompt (Continued) ---}, 
  }
}

\tcbset{
  promptinstruction/.style={
    colback=yellow!10!white, 
    colframe=finalblue, 
    coltitle=black,
    fonttitle=\bfseries\large\sffamily, 
    boxrule=1.5pt, 
    sharp corners, 
    left=8pt,
    right=8pt,
    top=8pt,
    bottom=8pt,
    title after break={\centering --- Model Prompt (Continued) ---},
  }
}

\tcbset{
  takeawaybox/.style={
    colback=white,
    colframe=finalblue,
    boxrule=1pt,
    arc=1pt,
    left=6pt,
    right=6pt,
    top=6pt,
    bottom=6pt,
    fonttitle=\bfseries,
  }
}

\newcommand{\affilA}{$^{1}$}       
\newcommand{\affilB}{$^{2}$}
\newcommand{\affilC}{$^{3}$}  
\newcommand{\affilD}{$^{4}$}   
\newcommand{\affilE}{$^{5}$}

\title{$\mathbf{\textsc{Semma}}$: A Semantic Aware Knowledge Graph Foundation Model}

\author{%
  Arvindh Arun\thanks{Core Contributors.} \affilA, Sumit Kumar$^{\circ}$ \affilB, Mojtaba Nayyeri \affilA,  Bo Xiong \affilC \\
   \textbf{Ponnurangam Kumaraguru} \affilB\textbf{,} \textbf{Antonio Vergari} \affilD\textbf{,} \textbf{Steffen Staab} \affilA$^{ }$\affilE
   \\\\
   \affilA Institute for AI, University of Stuttgart, \affilB IIIT Hyderabad, \\\affilC Stanford University, \affilD  University of Edinburgh, \affilE  University of Southampton\\
  \texttt{arvindh.arun@ki.uni-stuttgart.de}\\
}

\begin{document}
\maketitle

\begin{abstract}
Knowledge Graph Foundation Models (KGFMs) have shown promise in enabling zero-shot reasoning over unseen graphs by learning transferable patterns. However, most existing KGFMs rely solely on graph structure, overlooking the rich semantic signals encoded in textual attributes. We introduce $\ours$, a dual-module KGFM that systematically integrates transferable textual semantics alongside structure. $\ours$ leverages Large Language Models (LLMs) to enrich relation identifiers, generating semantic embeddings that subsequently form a textual relation graph, which is fused with the structural component. Across 54 diverse KGs, $\ours$ outperforms purely structural baselines like $\ultra$ in fully inductive link prediction. Crucially, we show that in more challenging generalization settings, where the test-time relation vocabulary is entirely unseen, structural methods collapse while $\ours$ is $2$x more effective. Our findings demonstrate that textual semantics are critical for generalization in settings where structure alone fails, highlighting the need for foundation models that unify structural and linguistic signals in knowledge reasoning.
\end{abstract}

\section{Introduction}
\label{sec:intro}
Knowledge Graphs (KGs) are used to store data and knowledge as triples \textit{(subject entity, relation, object entity)}, which form graphs \cite{Hogan_2021}. Knowledge can be expressed in its ontologies that enable logical reasoning over data. Knowledge graph embedding methods add reasoning by similarity and analogy, allowing for, e.g., recommending relationships between entities that have not been asserted explicitly and cannot be deduced deductively \cite{xiong-survey}. Such link prediction has been applied to recommender systems \cite{kg4rs}, entity linking \cite{kolitsas-etal-2018-end}, and question answering over knowledge graphs \cite{perevalov-etal-2022-knowledge}.

Analogous to Large Language Models (LLMs) that learn complex correlations between tokens from a large corpus and apply them in previously unseen contexts \cite{zhao2025surveylargelanguagemodels}, related work has started to investigate Knowledge Graph Foundation Models (KGFMs) to learn complex reasoning capabilities from \emph{training knowledge graphs} and apply them to previously unseen \emph{test knowledge graphs} \cite{galkin2024foundationmodelsknowledgegraph, zhang2025trixexpressivemodelzeroshot, huang2025expressiveknowledgegraphfoundation}.

These KGFMs exhibit several important advantages \cite{gfmsarehere} like \emph{(i) Broad applicability:} Zero-shot reasoning can be performed on unseen knowledge graphs. \emph{(ii) Efficiency:} While the training of knowledge-graph embedding methods is computationally highly challenging, performing a single round of inferences with zero-shot reasoning scales well to huge knowledge graphs. \emph{(iii) Effectiveness:} Reasoning patterns can be learned from rich knowledge graphs and applied to knowledge graphs that lack a rich ontology and densely linked entities.

Knowledge graph triples are expressed using symbols that denote entities and relations. While logical and similarity-based reasoning do not require human-understandable symbols, \citet{de2016names} have shown that symbols are ``often constructed mnemonically, in order to be meaningful to a human interpreter’’ and have established for over 500k datasets that there is significant mutual information between the symbols and their formal meaning.

Though related work has used word embeddings of knowledge graph symbols and other attributed texts (e.g.\ comments) when embedding individual knowledge graphs \cite{DBLP:conf/semweb/NayyeriWAARLS23, Yuan_Zhou_Chen_Wang_Liang_Liu_Huang_2025, kgbert, kicgpt, xu-etal-2021-fusing}, KGFMs have so far neglected the capabilities of symbols and other text attributions to generalize across knowledge graphs.

\begin{figure*}[ht]
    \centering
    \includegraphics[width=\textwidth]{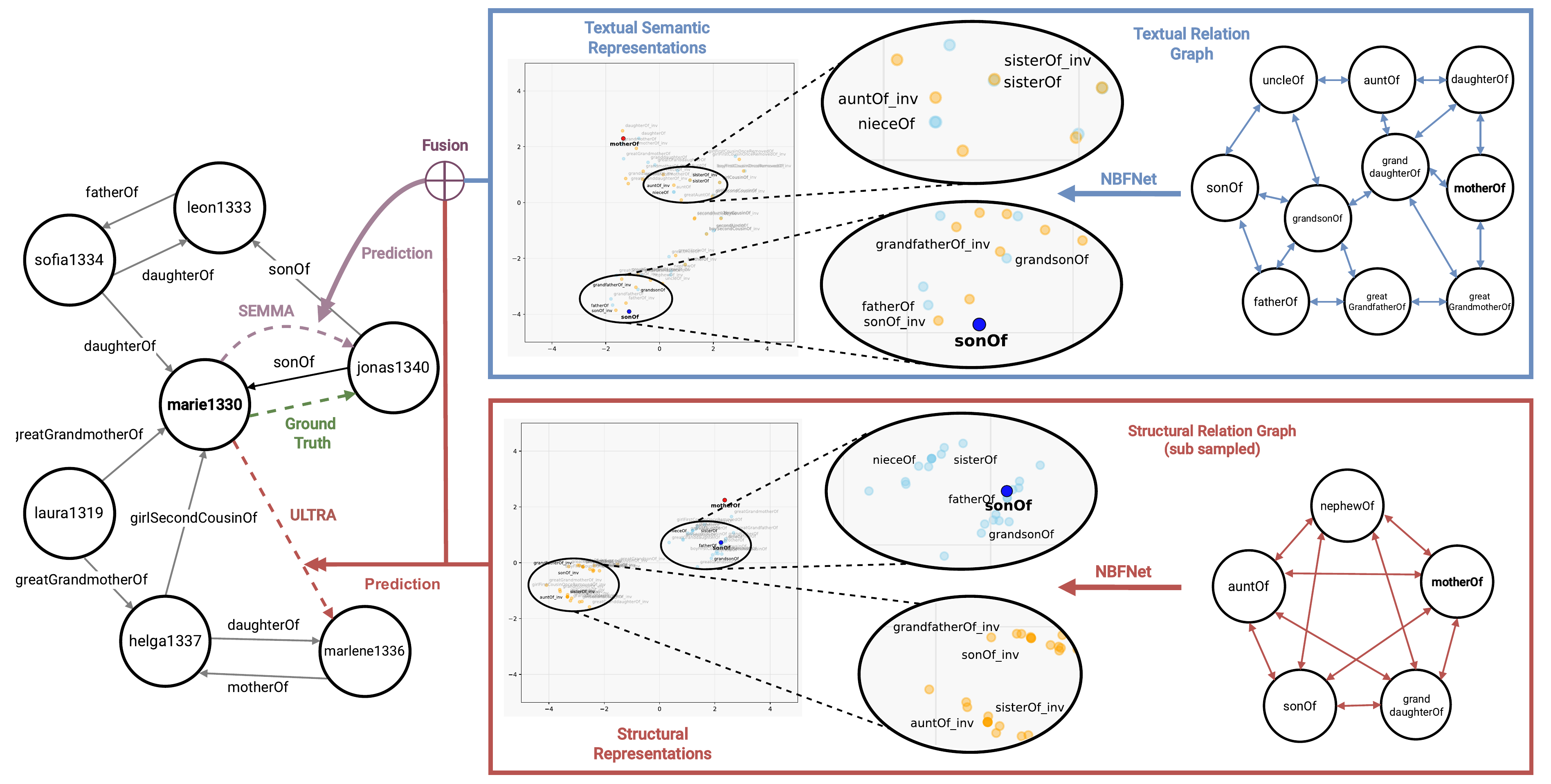}
    \caption{\textbf{$\ours$'s advantage in link prediction.} 
    \textbf{(Left)} $\ours$ correctly predicts \textit{(marie1330, motherOf, jonas1340)} from Metafam \citep{metafam} where $\ultra$ fails. Blue nodes correspond to relations, and orange nodes to their inverses. \textbf{(Right)} The Textual Relation Graph (top) is more ontologically coherent than the Structural Relation Graph (bottom), which is a clique, leading to different embeddings. \textbf{(Middle)} Textual semantic embeddings (top) show \textit{fatherOf} is near \textit{sonOf\_inv} (equivalence) and \textit{sisterOf} overlaps with \textit{sisterOf\_inv} (symmetry and inversion leading to equivalence), reflecting semantic understanding. Meanwhile, structural embeddings lack this clear organization for these pairs.
    }
    \label{fig:overview}
\end{figure*}

Generalizing within knowledge graph embedding methods to transfer embeddings from one or multiple knowledge graphs with millions of nodes to an unseen knowledge graph with largely distinct nodes can be hindered by a low signal-to-noise ratio \cite{truste}. However, we observe that knowledge graphs tend to have a huge number of nodes, but a much smaller number of highly informative relations. For example, Wikidata \cite{DBLP:journals/cacm/VrandecicK14} contains over 110 million nodes but only 12,681 relations as of May 19, 2025. This disparity indicates that the compact vocabulary of relations, rather than the vast and often graph-specific set of entities, provides a more robust transferable signal for generalization.

Therefore, we introduce $\ours$, which extends the capability of KGFMs to learn foundational knowledge graph semantics from the training knowledge graphs and apply it for reasoning on test knowledge graphs by analyzing (i) graph patterns \textbf{and} (ii) word embeddings of relation symbols. 

As illustrated in \Cref{fig:overview}, our approach extends $\ultra$ \citep{galkin2024foundationmodelsknowledgegraph} by constructing: (i) a \emph{structural relation graph}, which represents patterns that generalize from graph structures of the training graphs, similar to $\ultra$ , and (ii) a \emph{textual relation graph}, which represents patterns that are found by prompting an LLM to generate semantically rich descriptions for each relation identifier and/or attributed labels. Together, the two relation graphs allow for predicting high-quality links that would not be found by existing KGFM methods. We also make our codebase public. \footnote{\href{https://github.com/arvindh75/semma}{https://github.com/arvindh75/semma}} In summary, our contributions are:

\begin{itemize}
    \item We introduce $\ours$, a novel KGFM designed to perform zero-shot link prediction by learning from graph structures and the word embeddings of relation identifiers (and/or related text attributions).
    \item Extensive experiments on 54 diverse knowledge graphs demonstrating $\ours$'s superior performance over  KGFM baselines like $\ultra$ in fully inductive link prediction. Importantly, after identifying and mitigating data leakages, $\ours$'s advantage remains, highlighting its robust generalization capabilities.
\end{itemize}

\begin{figure*}[ht]
    \centering
    \includegraphics[width=\textwidth]{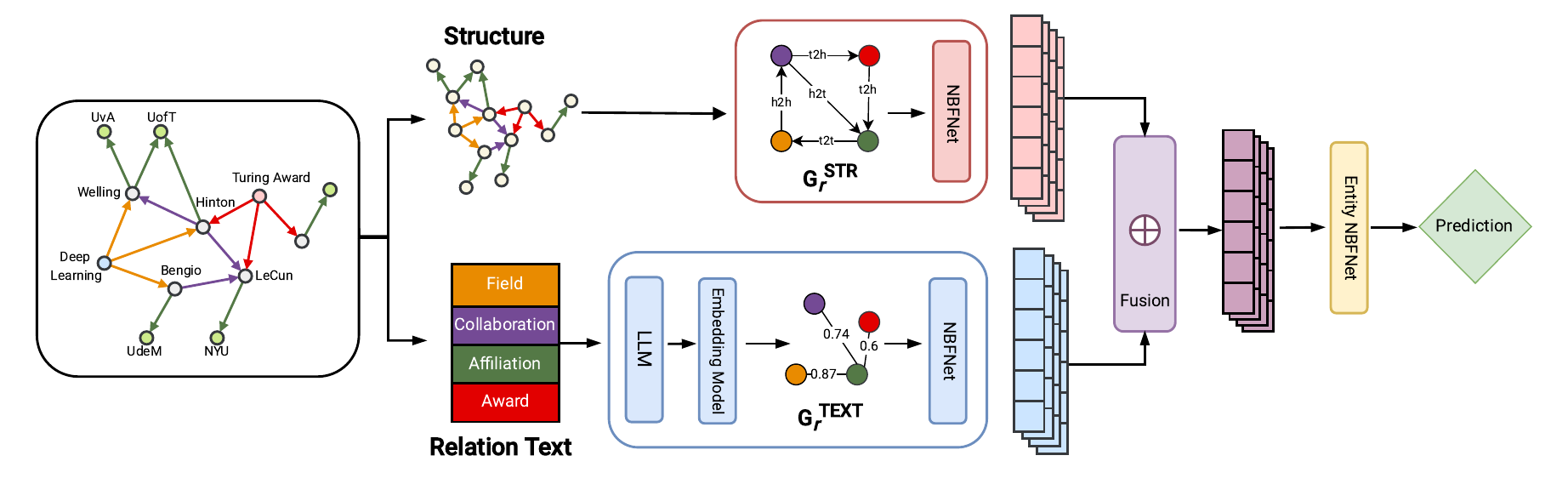}
    \caption{\textbf{Parallel Architecture of $\ours$.} 
    The \textit{structure processing module} (red) utilizes $\rgone$ and NBFNet to derive structure-based representations, similar to $\ultra$. Concurrently, the \textit{text processing module} (blue) leverages LLM enrichment of relation text and $\rgtwo$ with its own NBFNet to produce textual semantic representations. These two representations are fused and fed into an entity-level NBFNet to perform link prediction based on the input query.}
    \label{fig:architecture}
\end{figure*}

\section{Preliminaries}
\label{sec:preliminaries}
\textbf{Knowledge Graphs.} KGs represent knowledge in a directed edge-labeled graph.

\begin{definition}[Knowledge Graphs]
A Knowledge Graph $\mathcal{G}$ is a structure $\mathcal{G} = (\mathcal{E}, \mathcal{R}, \mathcal{F})$,
where $\mathcal{E}$ is a set of entities, $\mathcal{R}$ is a set of relation identifiers, and $\mathcal{F} \subseteq \mathcal{E} \times \mathcal{R} \times \mathcal{E}$ is a set of facts. 
\end{definition}
A fact $(h, r, t)\in \mathcal{F}$ (also referred to as a \emph{triple}) connects its head entity $h$ with its tail entity $t$ via the relation identifier $r$.

\noindent \textbf{Link prediction.} We aim at link prediction, a fundamental task for KG reasoning. We assume that one knowledge graph 
$\Gtruth=(\Etruth,\Rtruth,\Ftruth)$
 describes the set of facts that are true, another knowledge graph 
 $\Gobs=(\Eobs,\Robs,\Fobs)$ represents the set of facts that have been observed and are known to be true, and $\Eobs\subseteq\Etruth, \Robs\subseteq\Rtruth, \Fobs\subseteq\Ftruth$.

Given the observed graph $\Gobs$, link prediction aims to infer missing true triples $\Ftruth \setminus \Fobs$ using triples with either the tail or head entity masked (called queries), denoted as $(h, r, ?)$ and $(?, r, t)$.  The latter can equivalently be written as $(t,r^{-1}, ?)$, where $r^{-1}$ denotes the inverse relation of $r$. 
For learning, we split the observed KG into a training graph 
$\Gtrain=(\Etrain,\Rtrain,\Ftrain)$ and a test graph 
$\Gtest=(\Etest,\Rtest,\Ftest)$, such that 
$\Gtrain \cup \Gtest = \Gobs$.
The task falls into three regimes, based on the overlap between $\Gtrain$ and $\Gtest$:

\begin{itemize}[leftmargin=*,topsep=0pt,itemsep=0pt]
    \item \textbf{Transductive}. In this regime, the training graph is identical to the test graph ($\Gtrain = \Gtest$).
    Traditional KG embedding models like TransE \cite{TransE}, RotatE \cite{RotatE}, and ComplEx \cite{compleX}  learn embeddings for all entities and relations under this assumption. These models cannot predict links that involve entities or relations not observed and used for generating the embedding.
    \item \textbf{Partially Inductive.} In this regime, all relation identifiers are known at training time ($\Rtrain=\Rtest$), but not all the entities ($\Etrain\subset\Etest$). Inductive Graph Neural Networks (GNNs) or rule mining methods like NBFNet \cite{NBFNet} and GraIL \cite{grail} learn to generalize in order to predict links involving entities from $\Etest \setminus \Etrain$.  However, they cannot predict links involving unseen relation identifiers.
    \item \textbf{Fully Inductive.} In this most general regime, neither entities nor relations are fully known during training ($\Etrain\subset\Etest$ and $\Rtrain \subset \Rtest$). This regime best reflects the link prediction capabilities expected from a foundation model, namely, the ability to apply its pre-trained knowledge to entirely new, previously unseen KGs at test time. KGFMs, such as $\ultra$ \citep{galkin2024foundationmodelsknowledgegraph}, $\trix$ \citep{zhang2025trixexpressivemodelzeroshot} and $\mgnn$ \citep{huang2025expressiveknowledgegraphfoundation} learn widely transferable structural patterns. 
\end{itemize}

\noindent \textbf{Text-Attributed Knowledge Graphs.} 
Virtually all real-world knowledge graphs associate textual attributions with entities and relation identifiers.
\begin{definition}[Text-Attributed Knowledge Graphs]
    A Text-Attributed Knowledge Graph (TAKG) is a structure $\mathcal{G} = (\mathcal{E}, \mathcal{R}, \labl, \mathcal{F})$, with a knowledge graph $(\mathcal{E}, \mathcal{R}, \mathcal{F})$ and a labeling function $\labl:\EE\cup\RR\rightarrow\txt$, where $\txt$ denotes strings of arbitrary finite length over an alphabet $\Sigma$. Let $\mathcal{T_E} = \{\labl(e)\ |\ e \in \mathcal{E}\}$ and $\mathcal{T_R} = \{\labl(r)\ |\ r \in \mathcal{R}\}$ be the set of entity and relation labels.
\end{definition}

For example, Wikidata \cite{DBLP:journals/cacm/VrandecicK14} contains the triple \verb|(Q42 P27 Q145)|, with $\labl(\verb|Q42|) = \text{``Douglas Adams''},\ \labl(\verb|P27|) = \text{``country of citizenship''},\ \text{and}\ \labl(\verb|Q145|) = \text{``United Kingdom''}$. All of the 57 datasets from the benchmark for this task (see \Cref{sec:exp-setup}) attribute textual information. While some real-world TAKGs, like Wikidata, exhibit even more complex graph structures (e.g., involving time) and text attributes (e.g., multi-lingual), our definitions allow for a sufficiently expressive investigation of link prediction models in all of them.

To the best of our knowledge, the state-of-the-art in link prediction has either considered textual attributions but ignored the fully inductive regime \cite{Yuan_Zhou_Chen_Wang_Liang_Liu_Huang_2025, kgbert, kicgpt, xu-etal-2021-fusing}, or it has considered the fully inductive regime but has ignored text information \cite{ingram, NBFNet, grail}. Prior works like BLP \cite{Daza_2021}, StatiK \cite{markowitz-etal-2022-statik}, and SimKGC \cite{wang-etal-2022-simkgc} provide early empirical evidence that leveraging textual descriptions of entities and relations allows us to learn transferable representations for inductive prediction tasks without additional training.
Building on this, our main hypothesis is that leveraging textual attributes can enhance fully inductive link prediction. The closest work to ours is $\prolink$ \cite{prolink}, focused on working in the low-resource setting where the query relations occur sparsely in $\Gobs$. $\prolink$ prompts LLMs to predict potential entity connections for unseen relations during testing, aiming to mitigate data sparsity. However, both their motivation and final evaluation setup differ from ours.

\section{Foundation Models for KGs}
\label{sec:fmkg}
Unlike traditional KGE models trained on a single graph, a KG foundation model is pre-trained on a large and diverse collection of KGs, collected in $\Gtrain$. The objective is to learn transferable patterns that enable zero-shot generalization to new, unseen KGs at inference time without requiring fine-tuning on $\Gtest$.

\noindent \textbf{$\ultra$.} $\ultra$ \citep{galkin2024foundationmodelsknowledgegraph} is a KGFM designed for fully inductive link prediction, capable of zero-shot generalization to unseen KGs. However, it does not exploit textual attributions. $\ultra$ learns transferable structural patterns based on how relations interact within a graph, but it does not learn embeddings for specific relation types. Theoretically grounded by the principle of double equivariance \cite{double_equivariance}, $\ultra$ lifts the training and test graph $\Gobs$ to a higher-level structure, which we refer to as $\rgone$.
In $\rgone$, each node corresponds to a unique relation type or its inverse from $\Gobs$. Edges in $\rgone$ capture fundamental structural interactions between pairs of relations in $\Gobs$, categorized into head-to-head (h2h), tail-to-head (t2h), head-to-tail (h2t), tail-to-tail (t2t). $\rgone$ represents the transferable structural aspect, independent of the specific entity or relation vocabulary. $\ultra$ utilizes a GNN, specifically NBFNet \citep{NBFNet}, over $\rgone$, leveraging a labeling trick \cite{labeling-trick} to obtain conditional relation representations.
These conditional relation representations are used as input features for a second NBFNet operating on the original graph $\Gobs$ to perform the final link prediction. By learning relative structural patterns via $\rgone$ and conditioning representations on the query, $\ultra$ avoids learning fixed relation embeddings, enabling its zero-shot transfer capabilities. $\trix$ \citep{zhang2025trixexpressivemodelzeroshot} and $\mgnn$ \citep{huang2025expressiveknowledgegraphfoundation} are follow-up works that enhanced $\ultra$'s representation power by leveraging extra memory and compute.

\noindent \textbf{What do they lack?}
A key limitation of current KGFMs like $\ultra$, as discussed in \Cref{sec:intro}, is their sole reliance on graph structure, often neglecting or underutilizing the textual attributions commonly associated with entities and relations in most real-world KGs. These textual attributions, especially the semantics encoded in relation names or descriptions, can be crucial for accurate and generalizable reasoning but are often overlooked by structure-focused methods \citep{cornell-etal-2022-challenging, alam2024semanticallyenrichedembeddingsknowledge}. The generalization capabilities of LLMs have been widely recognized and leveraged in other domains \cite{li-etal-2024-fundamental}, yet their potential to enrich KGFMs by interpreting textual attributes remains largely unexplored.

\begin{table*}[ht]
\small
\centering
\begin{tabular}{l c c c c c c c || c c}
\toprule
 & & \multicolumn{2}{c}{\textbf{Inductive $e, r$}} & \multicolumn{2}{c}{\textbf{Inductive $e$}} & \multicolumn{2}{c}{\textbf{Transductive}} & \multicolumn{2}{c}{\textbf{Total Avg}}\\

\textbf{Model} & \textbf{Variant} & \multicolumn{2}{c}{(23 graphs)} & \multicolumn{2}{c}{(18 graphs)} & \multicolumn{2}{c}{(13 graphs)} & \multicolumn{2}{c}{(54 graphs)}\\
\cmidrule{3-10}
 & & \textbf{MRR} & \textbf{H@10} & \textbf{MRR} & \textbf{H@10} & \textbf{MRR} & \textbf{H@10} & \textbf{MRR} & \textbf{H@10} \\
\midrule
$\ultra$                        &  & $0.344$ & $0.511$ & $0.428$ & $0.570$ & $\bm{0.316}$ & $0.464$ & $0.365_{\pm 0.003}$ & $0.519_{\pm 0.004}$\\
$\ours$                       & & $\bm{0.350}$ & $\bm{0.514}$ & $\bm{0.447}$ & $\bm{0.584}$ & $\bm{0.316}$ & $\bm{0.467}$ & $\bm{0.374_{\pm 0.003}}$ & $\bm{0.526_{\pm 0.004}}$\\
\midrule
$\ours$ & $\textsc{Hybrid}$ & $0.353$ & $0.519$ & $0.448$ & $0.585$ & $0.318$ & $0.470$ & $0.376_{\pm 0.003}$ & $0.529_{\pm 0.003}$\\
\bottomrule
\end{tabular}
\caption{\textbf{Zero-shot results of $\ours$.} Zero-shot link prediction MRR and Hits@10 reported over 54 KGs averaged over 5 runs. $\ours$ outperforms $\ultra$ by considerable margins and $\ours\ \textsc{hybrid}$ increases the gap further.}
\label{tab:zeroshot}
\end{table*}



\section{\texorpdfstring{$\ours$}{OURS}: Integrating Structure and Textual Semantics}
\label{sec:method}
\noindent \textbf{What do we want from a KGFM for TAKGs?} 
This gap motivates the need for KGFMs specifically designed for TAKGs. In response, we propose $\ours$ and establish the following desiderata:\\
\circled{1} \textbf{Explicitly Leverage Textual Attributions.} The model should effectively utilize the 
 textual attributions present in KGs, when it is available.
\circled{2} \textbf{Maintain Robustness.} When textual attributions are not available, are noisy, or provide little semantic value, the model's performance should gracefully degrade to be at least as good as strong structure-only baselines like $\ultra$.

\noindent \textbf{Transferability of Ontological information.}
There are some key considerations to keep in mind while using the textual attributions in TAKGs,
\begin{enumerate}[leftmargin=*,topsep=0pt,itemsep=0pt]
    \item The usefulness of text attributions for link prediction varies across different datasets and even within a single dataset.
 Relation names in $\mathcal{T_R}$ might range from highly descriptive natural language phrases (e.g., \textit{country of citizenship}) to database paths (e.g., \textit{/film/film/genre}). Consequently, the semantic value inferable from this text can vary significantly.
    \item While entity text $(\mathcal{T_E})$ may sometimes be limited to identifiers or obfuscated for privacy, relations $(\mathcal{R})$ fundamentally require some semantic basis to be meaningful within a knowledge graph. Hence, we constrain our approach to only leverage the relation text/labels $(\mathcal{T_R})$. We assume that some textual representation for relations $(\mathcal{T_R})$ generally exists, even if it's merely an ID. 
\end{enumerate} 
\textbf{Our approach.} To address the established desiderata, we propose $\ours$ (\Cref{fig:architecture}), which explicitly separates structural processing (as in $\ultra$) from a module dedicated to processing relations' textual semantics derived from LLMs in parallel. This dual-module approach provides modularity, allowing the text processing module to be deactivated when the textual attributions are noisy or unhelpful, thus maintaining robust performance (\circled{2}\hspace{-1mm}). When suitable, the modules merge to effectively leverage structural and textual modalities (\circled{1}\hspace{-1mm}).

We modularize our pipeline into three independent components: utilizing LLMs to generate semantically rich text for each relation (\Cref{sec:extract-semantic-llms}), embedding the extracted text attributions into a vector space (\Cref{sec:extract-embeddings}), and using the embeddings to incorporate the semantics of text attributions into the prediction pipeline (\Cref{sec:use-embeddings}).

\subsection{Extracting textual semantic information from LLMs}
\label{sec:extract-semantic-llms}
The text processing module utilizes LLMs to process raw relation labels ($\mathcal{T_R}$) and generate richer textual representations suitable for subsequent embedding.
We leverage the inherent world knowledge and zero-shot capability of LLMs \cite{li2024zero,petroni2019language} to interpret relation semantics even for previously unseen KGs.

\noindent \textbf{What information can we extract from LLMs?}
For each relation identifier in $\mathcal{T_R}$, we prompt an LLM to generate two outputs:
\begin{enumerate}[leftmargin=*,topsep=0pt,itemsep=0pt]
    \item \textbf{Cleaned Relation Name.} A cleaned name without any special characters for semantically accurate tokenization \cite{kudo-richardson-2018-sentencepiece}. E.g., \textit{greatGranddaughterOf} to \textit{Great Granddaughter Of}.
    \item \textbf{Concise Textual Description.} A short definition explaining the core meaning or function of the relation and its inverse \citep{zrllm}. E.g., \textit{greatGranddaughterOf} to \textit{Female great-grandchild of}, providing deeper contextual understanding.
\end{enumerate}

\noindent \textbf{How do we extract this information from LLMs?}
We use a zero-shot prompt that provides the LLM with the relation's identifier, example triples from the KG for context, and strict output constraints. This method ensures the generation of consistent and semantically meaningful text suitable for embedding. Note that we only assume LLM outputs are \textit{beneficial} but not necessarily \textit{optimal} in all contexts. More details about the prompt are provided in \Cref{appendix:prompt}.

\subsection{Extracting embeddings from LLMs}
\label{sec:extract-embeddings}
Once we have the enriched textual representations from \Cref{sec:extract-semantic-llms}, we encode them into vector embeddings using transformers that capture their semantic meaning. These embeddings are the primary input for the text processing module.

\noindent \textbf{Which text representation yields the most suitable embeddings?} The process in Section 4.1 yields multiple textual candidates for each relation:
(1) the original identifier (\nollm),
(2) the LLM-generated cleaned name (\cleaned), and
(3) the LLM-generated textual description (\desc). Backed by the theory of concepts in the representation space \citep{park2025the}, we also create \cs, where the embeddings of all the above are combined through vector addition, and \cavg, where they are averaged in the vector space. We denote the text embedding of a relation $r$ by $\mathbf{\tau}_r$, which can be obtained by one of the above-mentioned approaches. 
We evaluate $\ours$ with all of the mentioned five variants, with more details in \Cref{sec:rq2}.

\noindent \textbf{How do we embed inverse relations}? For each relation $r$, $\ultra$ generates its virtual inverse relation, denoted by $r^{-1}$, without its ground-truth text. 
Since we cannot embed the inverse relations without text (except in \desc), we require a method to derive their semantic embeddings. 
Intuitively, if the original relation embedding signifies a transformation in a certain conceptual direction in the embedding space, its inverse should logically point the other way \cite{RotatE}. So, given a relation's embedding $\mathbf{\tau}_{r}$, we generate the inverse relation's embedding $\mathbf{\tau}_{r^{-1}}$ by rotating the original vector by 180 degrees to address this, i.e., $\mathbf{\tau}_{r^{-1}} = (\mathbf{I} - 2\frac{\mathbf{\tau}_{r} \mathbf{\tau}_{r}^{T}}{\|\mathbf{\tau}_{r}\|^2}) \mathbf{\tau}_{r}$ which simplifies to $ \mathbf{\tau}_{r^{-1}} = -\mathbf{\tau}_{r}$. While theoretically imperfect for symmetric relations (which we observe to be rare in practice), this method is a reasonable heuristic and proves empirically effective (seen in \Cref{tab:ablations}).

\subsection{Utilizing the embeddings}
\label{sec:use-embeddings}
The final stage integrates the textual relation embeddings (from \Cref{sec:extract-embeddings}) with the structural information processed by $\ours$ in parallel.

\noindent \textbf{How do we represent inter-relation semantics?} 
We construct a Textual Relation Graph ($\rgtwo$) where nodes are relations and weighted edges (with weights $\omega$) represent the cosine similarity between their textual embeddings, to explicitly model semantic proximity between relations. An edge in $\rgtwo$ is denoted by a triple $(u, \omega, v)$ where $u,v$ are two relations with $\omega = \cos(\tau_u,\tau_v)$. 
For efficiency and to avoid oversmoothing of representations, we filter the edges $\rgtwo$ based on edge weights ($\omega$) using sampling methods like choosing the Top-x\% and thresholding \cite{threshold}.

\noindent \textbf{How are textual and structural information processed?} 
We utilize NBFNet \cite{NBFNet} to process both the structural relation graph $\rgone$ and the textual relation graph $\rgtwo$ in parallel. 
Importantly, the text processing module leverages NBFNet's support for weighted edges, using the cosine similarities ($w$) in $\rgtwo$ to modulate message passing based on semantic relatedness. 

\begin{table*}[t]
\small
\centering
\begin{tabular}{l c c c c c c c || c c}
\toprule
 & & \multicolumn{2}{c}{\textbf{Inductive $e, r$}} & \multicolumn{2}{c}{\textbf{Inductive $e$}} & \multicolumn{2}{c}{\textbf{Transductive}} & \multicolumn{2}{c}{\textbf{Total Avg}}\\

\textbf{Model} & \textbf{Variant} & \multicolumn{2}{c}{(8 graphs)} & \multicolumn{2}{c}{(7 graphs)} & \multicolumn{2}{c}{(7 graphs)} & \multicolumn{2}{c}{(22 graphs)}\\
\cmidrule{3-10}
 & & \textbf{MRR} & \textbf{H@10} & \textbf{MRR} & \textbf{H@10} & \textbf{MRR} & \textbf{H@10} & \textbf{MRR} & \textbf{H@10} \\
\midrule
$\ultra$                        &  & $0.388$ & $0.585$ & $0.347$ & $0.475$ & $\bm{0.300}$ & $0.438$ & $0.347_{\pm 0.006}$ & $0.503_{\pm 0.010}$ \\
$\ours$                       & & $\bm{0.399}$ & $\bm{0.589}$ & $\bm{0.357}$ & $\bm{0.482}$ & $0.295$ & $\bm{0.441}$ & $\bm{0.353_{\pm 0.006}}$ & $\bm{0.508_{\pm 0.009}}$\\
\midrule
$\ours$ & $\textsc{Hybrid}$ & $0.406$ & $0.600$ & $0.357$ & $0.484$ & $0.298$ & $0.445$ & $0.356_{\pm 0.007}$ & $0.514_{\pm 0.010}$\\
\bottomrule
\end{tabular}
\caption{\textbf{Zero-shot results of $\ours$ on unleaked datasets.} Zero-shot link prediction MRR and Hits@10 reported over 22 KGs averaged over 5 runs after removing leaked datasets from testing. $\ours$ still outperforms $\ultra$.}
\label{tab:leaked-exp}
\end{table*}

Formally, let $\mathbf{h}_{u}^{(t)}, \mathbf{z}_{u}^{(t)}$ denote the state of relation node $u$ in $\rgone$ and $\rgtwo$ respectively at iteration $t$ (conditioned on query relation $r$). Let $\textsc{Upd}$, $\textsc{Msg}$, $\bigoplus$ denote the NBFNet update, message, and aggregation functions. Let $\mathbf{e}_{r'}$ denote the embedding of the edge type $r'\in \{h2h, t2h, h2t, t2t\}$ in $\rgone$. $\mathbf{h}_{u}^{(0)}$ is initialized as a $\mathbbm{1}_d$ for $r$ and ${0}_d$ for the rest of the relations. $\mathbf{z}_{u}^{(0)}$ are initialized with the respective embeddings after LLM enrichment, i.e., $\mathbf{z}_{u}^{(0)} = \tau_u$. Then the node embeddings are updated as follows,

\vspace{1em}

\scalebox{.86}{$
\begin{aligned}
    \mathbf{h}_{u}^{(t+1)} = \textsc{Upd} \left( \mathbf{h}_{u}^{(t)}, \bigoplus\nolimits_{\substack{v \in \mathcal{N}(u) \\ (u,r',v) \in \rgone}} \textsc{Msg}(\mathbf{h}_{v}^{(t)}, \mathbf{e}_{r'}) \right)
\end{aligned}
$}

\vspace{1em}

\scalebox{.89}{$
\begin{aligned}
    \mathbf{z}_{u}^{(t+1)} = \textsc{Upd} \left( \mathbf{z}_{u}^{(t)}, \bigoplus\nolimits_{\substack{v \in \mathcal{N}(u) \\ (u,\omega,v) \in \rgtwo}} \textsc{Msg}(\mathbf{z}_{v}^{(t)}, \mathbf{\omega}) \right)
\end{aligned}
$}

\vspace{1em}

\noindent \textbf{How are structural and textual signals fused?} The outputs from the parallel structural ($\rgone$) and textual ($\rgtwo$) NBFNet modules are combined to produce the final embeddings. 
We initially explored simple concatenation for merging the structural $(\mathbf{h}_{u})$ and textual signals $(\mathbf{z}_{u})$, but this more than doubled the model parameters to handle the extra dimensions and also resulted in significantly longer convergence times. Hence, we evaluate standard fusion techniques ($F$) like MLP and Attention to reduce them back to the original dimensions. Formally, let $n$ denote the final iteration,
\begin{equation*}
    \mathbf{H}_{u}^{(n)} = F(\mathbf{h}_{u}^{(n)} \oplus (\alpha \mathbf{z}_{u}^{(n)} + 0_d)), \, \alpha \in \{0,1\}
\end{equation*}
where $\oplus$ is the concatenation operator and $\alpha$ is a hyperparameter that disables the text processing module if set to zero ($\alpha = 1$ by default). This is then passed on to the Entity NBFNet for final prediction. We use the same training objective as $\ultra$ for the whole pipeline. The primary overhead introduced by $\ours$ is a one-time LLM enrichment step, which has a computational complexity of $O(|\Rtest|)$, scaling linearly with the number of relations in $\Gtest$. More details in \Cref{appendix:Implementation}.

\section{Experiments}
Through our experiments, we  address the following questions,
\begin{enumerate}[label=\textbf{RQ\arabic*:}, leftmargin=2.5em]
    \item Where does textual semantics matter?
    \item Does adding textual semantics increase average performance?
    \item Can textual semantics help generalize to newer, harder settings?
    \item Do the existing benchmarks suffice?
\end{enumerate}

\subsection{Experimental Setup}
\label{sec:exp-setup}
\noindent \textbf{Datasets, Baseline, and Setup.} We use the same setup as $\ultra$, pretraining on 3 datasets and testing on 54. They are categorized into three categories: Transductive, Partially Inductive (Inductive $e$), and Fully Inductive (Inductive $e, r$), depending on the overlap between $\Gtrain$ and $\Gtest$ as discussed in \Cref{sec:preliminaries}. We use $\ultra$ as the main baseline for fair comparison. We leave it for future work to extend this to $\trix$ and $\mgnn$, where our framework can be adapted to fit in their pipeline. $\ours$ is relatively small and has around 227k parameters. All experiments were run on 2 NVIDIA A100 GPUs. More details in \Cref{appendix:Datasets,appendix:Implementation}.

\noindent \textbf{Metrics.} We report the Mean Reciprocal Rank (\textbf{MRR}) and Hits@10 (\textbf{H@10}) averaged across the categories and over all the datasets. The reported values are averaged across 5 runs, along with the standard deviation values.

\noindent \textbf{Design Choices.} We use gpt-4o-2024-11-20 \cite{openai2024gpt4ocard} as the LLM, jina-embeddings-v3 \cite{jina-ai} as the text embedding model, \cs\ for combining the embeddings and a threshold of $0.8$ to construct $\rgtwo$ based on insights from \cite{threshold} and MLP for fusing $\rgone$ and $\rgtwo$. Ablations of these choices are discussed in \Cref{sec:rq2}.

\begin{table}[]
\small
\centering
\begin{tabular}{l c c}
\toprule
\textbf{Dataset} & \textbf{MRR} & \textbf{H@10}\\
\midrule
Metafam             & $+0.3303$ & $+0.2935$\\
WN18RRInductive:v1  & $+0.1058$ & $+0.0550$\\
ConceptNet100k      & $+0.0620$ & $+0.1188$\\
\midrule
YAGO310             & $-0.0879$  & $-0.0626$\\        
Hetionet            & $-0.0556$  & $-0.0487$\\     
WikiTopicsMT4:sci   & $-0.0364$  & $-0.0709$\\
\bottomrule
\end{tabular}
\caption{\textbf{Maximum and Minimum improvements of $\ours$.} $\ours$ has significant increases on some datasets while there are drops in a few others.}
\label{tab:max-min-diff}
\end{table}

\begin{table*}[th]
\small
\centering
\begin{tabular}{l cc cc cc cc}
\toprule
\textbf{Dataset} 
& \multicolumn{2}{c}{\textbf{Split Statistics}}
& \multicolumn{2}{c}{\textbf{MRR}} 
& \multicolumn{2}{c}{\textbf{Hits@3}} 
& \multicolumn{2}{c}{\textbf{Hits@10}} \\
\cmidrule(lr){2-3} \cmidrule(lr){4-5} \cmidrule(lr){6-7} \cmidrule(lr){8-9}
& \textbf{$|\Ftest|$} & \textbf{$|\Ftruth \setminus \Fobs|$} 
& \textbf{\ultra} & \textbf{\ours}
& \textbf{\ultra} & \textbf{\ours}
& \textbf{\ultra} & \textbf{\ours} \\
\midrule
FBNELL  & 5750 & 335 & $0.025$ & $\bm{0.058}_{(\textcolor{igreen}{+128\%})}$ & $0.035$ & $\bm{0.067}_{(\textcolor{igreen}{+87\%})}$ & $0.064$ & $\bm{0.135}_{(\textcolor{igreen}{+112\%})}$\\
Metafam  & $5900$ & $177$ & $0.098$ & $\bm{0.180}_{(\textcolor{igreen}{+83\%})}$ & $0.079$ & $\bm{0.206}_{(\textcolor{igreen}{+161\%})}$ & $0.138$ & $\bm{0.338}_{(\textcolor{igreen}{+145\%})}$ \\
\bottomrule
\end{tabular}
\caption{\textbf{$\ours$ vs $\ultra$ evaluated on a harder setting with inductive relation vocabulary.} $\ours$, being more robust to new relation vocabulary, outperforms $\ultra$ by almost 2x on all the metrics.}
\label{tab:takeaway3}
\end{table*}

\subsection{Where does textual semantics matter?}
\label{sec:rq1}
Before looking at the average numbers across categories, we first analyze the impact of $\ours$ on the dataset level. From \Cref{tab:max-min-diff}, we can observe that for datasets like Metafam, the performance increase is almost 2x, while there are some datasets like YAGO310, where there is a considerable drop in MRR and H@10. We provide more insights on this in \Cref{appendix:performance_variance}. This hints at how textual semantics can be helpful for some datasets, subject to how rich the relation texts are. To leverage this insight, we introduce the $\ours\ \textsc{Hybrid}$ variant, where we switch off the text processing module based on the validation set's performance.

\subsection{Does adding textual semantics increase average performance?}
\label{sec:rq2}
From \Cref{tab:zeroshot}, we can see that $\ours$ clearly outperforms $\ultra$ averaged across all 54 datasets, with both MRR and H@10 having considerable improvements.
Note that in the domain of KGFMs, increases of this magnitude are significant, as the tasks are quite demanding \cite{zhang2025trixexpressivemodelzeroshot, huang2025expressiveknowledgegraphfoundation}. We also perform the Mann-Whitney U test to confirm the statistical significance of the performance gains of $\ours$. Total Avg MRR has a p-value of 0.0040 (< 0.05) across the runs, and Total Avg Hits@10 has a p-value of 0.0278 (< 0.05). We provide dataset-wise results in \Cref{appendix:fullresults}.

\begin{table}[h]
\centering
\scalebox{.9}{
\begin{tabular}{c r c c}
\toprule
& \textbf{Design choices} & \multicolumn{2}{c}{\textbf{Total Avg}} \\
&  & \multicolumn{2}{c}{(54 graphs)}\\
\cmidrule(lr){3-4} & & \textbf{MRR} & \textbf{H@10} \\
\midrule
\multirow{3}{*}{\rotatebox{90}{\textbf{LLMs}}} & gpt-4o-2024-11-20     & \bm{$0.377$} & $0.529$ \\
& deepseek-chat-v3-0324 & $0.375$ & \bm{$0.530$}\\
& qwen3-32b   & $0.368$     & $0.519$\\
\midrule
\multirow{2}{*}{\rotatebox{90}{\textbf{LM}}} & jina-embeddings-v3     & \bm{$0.377$} & \bm{$0.529$} \\
& Sentence-BERT & $0.368$ & $0.528$ \\
\midrule
\multirow{2}{*}{\rotatebox{90}{\textbf{$F$}}} & MLP     & \bm{$0.377$} & \bm{$0.529$} \\
& Attention & $0.369$ & $0.525$ \\
\midrule
\multirow{5}{*}{\rotatebox{90}{\textbf{Text}}} & $\textsc{rel\_name}$     & $0.369$ & $0.528$ \\
& $\textsc{llm\_rel\_name}$    & $0.375$ & $0.529$ \\
& $\textsc{llm\_rel\_desc}$   &  $0.373$ & $0.524$ \\
& \cs    & \bm{$0.377$} & \bm{$0.529$} \\
& \cavg & $0.374$ & $0.523$ \\
\midrule
\multirow{2}{*}{\raisebox{-0.3\height}{\rotatebox[origin=c]{90}{\textbf{$\rgtwo$}}}} & Threshold (0.8)     & \bm{$0.377$} & \bm{$0.529$} \\
& Top-x\% (20\%) & $0.371$ & $0.525$ \\
\bottomrule
\end{tabular}
}
\caption{\textbf{Ablations of Design Choices.} We conduct a rigorous search over settings for $\ours$.}
\label{tab:ablations}
\end{table}

\noindent \textbf{How robust is $\ours$ to design choices?} We ablate all the components of $\ours$ to study the impact of the design choices we made. We experiment with other cheaper open-weight LLMs: DeepSeek V3 \cite{deepseekai2025deepseekv3technicalreport} and Qwen3 32b \cite{yang2025qwen3technicalreport}, SentenceBERT \cite{sentBert} as the embedding model, and other choices described in \Cref{sec:use-embeddings} and \Cref{sec:extract-embeddings}. Results from \Cref{tab:ablations} reinforce our final design choices, with regime-wise results in \Cref{tab:LLM-full}.

\begin{tcolorbox}[takeawaybox]
\textbf{Takeaway 1.} Textual attributions can add value to link prediction compared to purely structural information. Therefore, $\ours$ can outperform $\ultra$ by a considerable margin.
\end{tcolorbox}

\subsection{Can textual semantics help generalize to newer, harder settings?}
\label{sec:rq3}
Similar to the low-resource setting proposed by \citet{prolink}, we create a harder evaluation setting where the relation vocabulary of the queries is disjoint from $\Rtest$, inspired by time-evolving KGs \citep{takeaway3} where new relations are potentially introduced over time. We discuss this setting in detail in \Cref{appendix:harder}. We adapt $\ultra$ and $\ours$ to work in this setting by generating a new $\rgone$ and $\rgtwo$ for each query. For queries that share the same head but have different relations, the corresponding $\rgone$ will be identical, causing $\ultra$ to produce the same predictions for all such queries, whereas $\rgtwo$ will be able to distinguish them based on the relation identifier. For a working example, refer to \Cref{fig:t3-example}. As observed in \Cref{tab:takeaway3}, $\ultra$'s performance drops substantially in the new setting where $\ours$ is 2x better.

\begin{tcolorbox}[takeawaybox]
\textbf{Takeaway 2.} The limitation of purely structural approaches like $\ultra$ becomes evident when query relations are disjoint from $\Rtest$, where $\ours$ still performs competently.
\end{tcolorbox}

\subsection{Do the existing benchmarks suffice?}
\label{sec:rq4}
In some datasets, we identified an overlap where triples from the pretraining data appeared in the test data. We call this common set of triples, the ``leaked set''. To assess the impact of this, we conduct a study where we remove all datasets exhibiting any leakage from our evaluation and report these ``unleaked'' results in \Cref{tab:leaked-exp}, with more dataset-wise leakage statistics reported in \Cref{appendix:Leakage}. From the results in \Cref{tab:leaked-exp}, $\ours$ maintains its superior performance over $\ultra$ on this cleaner, more challenging subset of 22 KGs. However, this analysis also highlights a broader issue: the current KGFM benchmarks, while diverse in some structural aspects, often draw from a limited set of popular, large-scale KGs (such as Freebase and WordNet) for both pretraining and testing. This homogeneity can lead to an overestimation of true generalization to genuinely novel domains and ontologies not represented in these common sources.

\begin{tcolorbox}[takeawaybox]
\textbf{Takeaway 3.} We need new benchmarks in the field that are from diverse domains and not derived from the same popular KGs to evaluate the actual effectiveness of KGFMs.
\end{tcolorbox}

\section{Conclusion and Future Directions}
We introduce $\ours$, a KGFM that utilizes LLM-derived textual semantics for relations alongside structure. By incorporating a textual relation graph, $\ours$ outperforms purely structural baselines such as $\ultra$ across 54 diverse KGs and shows strong generalization in harder settings with unseen relation vocabularies. This work is a first step toward semantically grounded KGFMs. While we focused on relation-level semantics, future works need to explore extending $\ours$ to entity-level text, richer multilingual inputs, and integration into more expressive models like $\trix$ and $\mgnn$. Building on our findings regarding current dataset limitations, the domain also needs new benchmarks with KGs from genuinely diverse and unseen domains to rigorously evaluate true generalization.
Finally, investigating how to extend $\ours$ to yield calibrated predictions \citep{NEURIPS2023_f4b76818}, 
and going beyond link prediction and answering a variety of complex queries \citep{DBLP:conf/nips/RenL20,galkin2024a,gregucci2025complex,he2025dage} are promising future research avenues. 

\section*{Limitations}
While $\ours$ demonstrates a promising direction for KGFMs, several limitations and avenues for future work remain. Firstly, our current approach exclusively focuses on relation text, leaving the semantic information often present in entity names and descriptions untapped; future iterations should explore integration of both. Secondly, the quality of the textual representations is inherently tied to the capabilities and outputs of the upstream LLM used for enrichment. While there are no direct risks of our work, there is a possibility of biases present in LLMs leaking into our pipeline. Next, while $\ours$ advances upon $\ultra$, future works should investigate its semantic enrichment pipeline with more recent, expressive KGFMs (such as $\trix$ or $\mgnn$). There is also scope to explore more adaptive fusion mechanisms that dynamically weigh structural versus textual signals, potentially improving upon the current $\ours\ \textsc{Hybrid}$'s validation-dependent switch, especially when relation text quality varies significantly. Finally, $\ours$ only explores ontological concepts that can be modeled with textual semantic similarity; there is scope for future work to broaden the horizon.

\section*{Acknowledgments}
The authors would like to thank Akshit Sinha for helpful feedback and help with the figures. AA was funded by the CHIPS Joint Undertaking (JU) under grant agreement No. 101140087 (SMARTY), and by the German Federal Ministry of Education and Research (BMBF) under the sub-project with the funding number 16MEE0444. 
AV was supported by the ``UNREAL: Unified Reasoning Layer for Trustworthy ML'' project (EP/Y023838/1) selected by the ERC and funded by UKRI EPSRC. MN acknowledges BMBF support through the ATLAS project (031L0304A).
AA thanks the International Max Planck Research School for Intelligent Systems (IMPRS-IS) and the European Laboratory for Learning and Intelligent Systems (ELLIS) PhD program for support. The authors gratefully acknowledge compute time on HoreKa HPC (NHR@KIT), funded by the BMBF and Baden-Württemberg’s MWK through the NHR program, with additional support from the DFG; and on the Artificial Intelligence Software Academy (AISA) cluster funded by the Ministry of Science, Research and Arts of Baden-Württemberg.

\section*{Author contributions}
AA conceived the project based on discussions with SS and led the overall design of $\ours$. AA led the setup and experiments of RQ1 (\Cref{sec:rq1}), RQ2 (\Cref{sec:rq2}) with the help of SK. AA and SK wrote the necessary code and ran experiments for all the RQs. AA and MN discovered the data leakage problem and, with the help of AV, proposed and ran experiments for RQ4 (\Cref{sec:rq4}) with help from SK. Based on an initial idea by BX, SK led the setup and experiments for RQ3 (\Cref{sec:rq3}) with the help of AA. AA wrote the initial draft with help from SK, where all other authors actively contributed to refining it. MN, BX, and AV actively advised on the design of all the experiments. SS and PK provided feedback and advice throughout the project.

\bibliography{main}

\newpage

\appendix
\section{Prompt}
\label{appendix:prompt}

We decompose the prompt into two steps - one to extract the cleaned relation names and the second to obtain the relation descriptions. We define clear rules for each of them and specify the output format. The exact prompt is listed in \Cref{fig:exact-prompt} and the system instruction in \Cref{fig:sys-ins}.

\begin{figure}[!h]
\begin{tcolorbox}[promptinstruction, title=System Instruction]
Provide exactly two separate JSON objects in your response, corresponding to each step, strictly in the order presented above. Do not include additional explanations or metadata beyond the specified JSON objects. Always provide both JSON objects and ensure they contain all the original relation names as keys.
\end{tcolorbox}
\caption{System Instruction used to ensure output consistency.}
\label{fig:sys-ins}
\end{figure}

We use the OpenRouter\footnote{https://openrouter.ai/} API to query the LLMs. In total, the experiments were pretty cost-effective, we spent \$8 on openai/gpt-4o-2024-11-20, \$0.8 on deepseek/deepseek-chat-v3-0324, and \$0.67 on qwen/qwen3-32b-04-28, totalling to less than \$10 for all the 57 datasets cumulatively. We also attach a sample output from openai/gpt-4o-2024-11-20, which can be found at \Cref{fig:gptout}.

\begin{figure*}[h]
\label{fig:ex-output}
\begin{tcolorbox}[promptinstruction, title=Sample Output]
\begin{lstlisting}[basicstyle=\small\ttfamily, showstringspaces=false, breaklines=true, columns=flexible]
"cleaned_relations": {
    "Causes": "Causes",
    "/organization/organization/headquarters./location/mailing_address/citytown": 
        "Headquarters City",
    "GpMF": "Gene participates Molecular Function"
    ...
}

"relation_descriptions": {
    "Causes": ["leads to effect", "effect caused by"],
    "/organization/organization/headquarters./location/mailing_address/citytown": 
        ["Headquarters located in city", "City has headquarters of"],
    "GpMF": ["Gene contributes to molecular function", "Molecular function involves gene"]
    ...
}
\end{lstlisting}
\end{tcolorbox}
\caption{Sample outputs from GPT 4o.}
\label{fig:gptout}
\end{figure*}

\begin{figure*}[h]
\centering
\scalebox{.9}{
\begin{tcolorbox}[promptinstruction, title=LLM Prompt for Relation Text Enrichment]
You will be provided with a list of relation names, each accompanied by exactly one example triple from a knowledge graph. Follow the instructions below carefully, strictly adhering to the output formats specified.\\

\textbf{Step 1:} \textit{Convert Relation Names to Human-Readable Form.}\\
Clean each provided relation name, converting it into plaintext, human-readable form.\\

Output Format (JSON Dictionary):
\{\{\\
"original\_relation\_name1": "Clean Human-Readable Form",\\
"original\_relation\_name2": "Clean Human-Readable Form",\\
...\\
\}\}\\

\textbf{Step 2:} \textit{Generate Short Descriptions}

For each provided relation, generate a concise description (3-4 words) that clearly captures its semantic meaning based on the given example triple as context. Also, for each relation, generate a description of its supposed inverse relation. These descriptions will be converted into embeddings using jinaai/jina-embeddings-v3 to uniquely identify relations and to measure semantic similarities. So, avoid using common or generic words excessively, and do NOT reuse other relation names, to prevent false semantic similarities. Follow the rules below,\\

\textit{Be Concise and Precise:} Use as few words as possible while clearly conveying the core meaning. Avoid filler words, unnecessary adjectives, and overly generic language.\\

\textit{Emphasize Key Semantics:} Focus on the distinctive action or relationship the relation name implies. Ensure that the description highlights the unique aspects that differentiate it from similar relations.\\

\textit{Handle Negation Carefully:} If the relation involves negation (e.g., ``is not part of''), state the negation explicitly and unambiguously. Ensure that the description for a negated relation is clearly distinguishable from its affirmative counterpart.\\

\textit{Avoid Common Stopwords as Filler:} Do not use common stopwords or phrases that add little semantic content. Every word should contribute meaning. Do not use repetitive words to avoid creating false semantic similarities.\\

\textit{Take care of symmetry:} Ensure that for relations that are symmetric, the description does not change for its inverse relation.\\

Output Format (JSON Dictionary):
\{\{\\
"original\_relation\_name1": ["concise description", "concise inverse relation description"],\\
"original\_relation\_name2": ["concise description", "concise inverse relation description"],\\
...\\
\}\}\\

List of Relations:\\
relation\_name: ``exist as'' ; example: (``chloride'', ``exist as'', ``crystal'')\\
relation\_name: ``concept:statehascapital'' ; example: (``concept:stateorprovince:mn'', ``concept:statehascapital'', ``concept:city:st\_paul'')\\
...
\end{tcolorbox}
}
\caption{LLM Prompt for relation text enrichment}
\label{fig:exact-prompt}
\end{figure*}

\section{Datasets}
\label{appendix:Datasets}

We follow the same evaluation strategy as $\ultra$ and conduct experiments on 57 publicly available KGs spanning a variety of sizes and domains. These datasets are grouped into three generalisation regimes: transductive, inductive with new entities ($e$), and inductive with new entities and relations ($e, r$) at test time. The detailed statistics for transductive datasets is presented in \Cref{tab:transductive}, inductive ($e$) in 
\Cref{tab:inductive_entity_datasets} and inductive ($e, r$) in \Cref{tab:inductive_entity_relation_datasets}. Consistent with $\ultra$, we perform tail-only evaluation on FB15k237\_10, FB15k237\_20, and FB15k237\_50, where predictions are restricted to the form $(h, r, ?)$ during test time.

\begin{table*}[ht]
\centering
\small
\begin{tabular}{l l r r r r r l}
\toprule
\textbf{Dataset} & \textbf{Reference} & \textbf{Entities} & \textbf{$|\mathcal{R}|$} & \textbf{Train} & \textbf{Valid} & \textbf{Test} & \textbf{Task} \\
\midrule
CoDExSmall       & \citet{codex}          & 2034   & 42    & 32888   & 1827   & 1828   & h/t \\
WDsinger          & \citet{sparseFB-WD-NELL}                 & 10282  & 135   & 16142   & 2163   & 2203   & h/t \\
FB15k237\_10      & \citet{sparseFB-WD-NELL}                 & 11512  & 237   & 27211   & 15624  & 18150  & tails \\
FB15k237\_20      & \citet{sparseFB-WD-NELL}                 & 13166  & 237   & 54423   & 16963  & 19776  & tails \\
FB15k237\_50      & \citet{sparseFB-WD-NELL}                 & 14149  & 237   & 136057  & 17449  & 20324  & tails \\
FB15k237          & \citet{Freebase}         & 14541  & 237   & 272115  & 17535  & 20466  & h/t \\
CoDExMedium      & \citet{codex}          & 17050  & 51    & 185584  & 10310  & 10311   & h/t \\
NELL23k           & \citet{sparseFB-WD-NELL}                 & 22925  & 200   & 25445   & 4961   & 4952   & h/t \\
WN18RR            & \citet{WN18RR}           & 40943  & 11    & 86835   & 3034   & 3134   & h/t \\
AristoV4          & \citet{Aristo}               & 44949  & 1605  & 242567  & 20000  & 20000  & h/t \\
Hetionet          & \citet{hetionet}        & 45158  & 24    & 2025177 & 112510 & 112510 & h/t \\
NELL995           & \citet{NELL995}              & 74536  & 200   & 149678  & 543   & 2818   & h/t \\
CoDExLarge       & \citet{codex}          & 77113  & 69    & 511359  & 30622  & 30622  & h/t \\
ConceptNet100k    & \citet{ConceptNet}           & 78334  & 34    & 100000  & 1200   & 1200   & h/t \\
DBpedia100k       & \citet{DBPedia}               & 99604  & 470   & 597572  & 50000   & 50000   & h/t \\
YAGO310           & \citet{Yago}       & 123182 & 37    & 1079040 & 5000   & 5000   & h/t \\
\bottomrule
\end{tabular}
\caption{Dataset-wise statistics for the 16 KGs evaluated under the transductive regime. \textbf{Entities} and \textbf{$|\mathcal{R}|$} indicate the vocabulary sizes of entities and relations, respectively. \textbf{Train}, \textbf{Valid}, and \textbf{Test} represent the number of triples in each split. The \textbf{Task} column specifies whether the model is evaluated on both head and tail prediction (\texttt{h/t}) or only on tail prediction (\texttt{tails}), consistent with $\ultra$'s evaluation setup.}
\label{tab:transductive}
\end{table*}

\begin{table*}[ht]
\centering
\small
\begin{tabular}{l | r | rr | rrr | rrr}
\toprule
\textbf{Dataset} & \textbf{$|\mathcal{R}|$}  
& \multicolumn{2}{c}{\textbf{Train Graph}} 
& \multicolumn{3}{c}{\textbf{Validation Graph}} 
& \multicolumn{3}{c}{\textbf{Test Graph}} \\
\cmidrule(lr){3-4} \cmidrule(lr){5-7} \cmidrule(lr){8-10}
 & & \textbf{$|\mathcal{E}|$} & \textbf{$|\mathcal{F}|$} 
   & \textbf{$|\mathcal{E}|$} & \textbf{$|\mathcal{F}|$} & \textbf{Test} 
   & \textbf{$|\mathcal{E}|$} & \textbf{$|\mathcal{F}|$} & \textbf{Test} \\
\midrule
FB v1      & 180 & 1594  & 4245   & 1594  & 4245   & 489   & 1093  & 1993   & 411 \\
FB v2      & 200 & 2608  & 9739   & 2608  & 9739   & 1166  & 1660  & 4145   & 947 \\
FB v3      & 215 & 3668  & 17986  & 3668  & 17986  & 2194  & 2501  & 7406   & 1731 \\
FB v4      & 219 & 4707  & 27203  & 4707  & 27203  & 3352  & 3051  & 11714  & 2840 \\
WN v1      & 9   & 2746  & 5410   & 2746  & 5410   & 630   & 922   & 1618   & 373 \\
WN v2      & 10  & 6954  & 15262  & 6954  & 15262  & 1838  & 2757  & 4011   & 852 \\
WN v3      & 11  & 12078 & 25901  & 12078 & 25901  & 3097  & 5084  & 6327   & 1143 \\
WN v4      & 9   & 3861  & 7940   & 3861  & 7940   & 934   & 7084  & 12334  & 2823 \\
NELL v1    & 14  & 3103  & 4687   & 3103  & 4687   & 414   & 225   & 833    & 201 \\
NELL v2    & 88  & 2564  & 8219   & 2564  & 8219   & 922   & 2086  & 4586   & 935 \\
NELL v3    & 142 & 4647  & 16393  & 4647  & 16393  & 1851  & 3566  & 8048   & 1620 \\
NELL v4    & 76  & 2092  & 7546   & 2092  & 7546   & 876   & 2795  & 7073   & 1447 \\
\midrule
ILPC Small  & 48  & 10230 & 78616  & 6653  & 20960  & 2908  & 6653  & 20960  & 2902 \\
ILPC Large  & 65  & 46626 & 202446 & 29246 & 77044  & 10179 & 29246 & 77044  & 10184 \\
\midrule
HM 1k     & 11  & 36237 & 93364  & 36311 & 93364  & 1771  & 9899  & 18638  & 476 \\
HM 3k     & 11  & 32118 & 71097  & 32250 & 71097  & 1201  & 19218 & 38285  & 1349 \\
HM 5k     & 11  & 28601 & 57601  & 28744 & 57601  & 900   & 23792 & 48425  & 2124 \\
\midrule
IndigoBM  & 229 & 12721 & 121601 & 12797 & 121601 & 14121 & 14775 & 250195 & 14904 \\
\bottomrule
\end{tabular}
\caption{Dataset-wise statistics for the 18 datasets used under the inductive entity ($e$) generalization regime. \textbf{$|\mathcal{R}|$} and \textbf{$|\mathcal{E}|$} indicate vocabulary sizes of relations and entities present in each split, respectively. \textbf{Train}, \textbf{Validation}, and \textbf{Test Graphs} include the number of entities and triples present in each split. The \textbf{Test} columns denote the number of link prediction queries evaluated in each corresponding graph. The first part of the datasets is from \citet{grail}, the second from \citet{ILPC}, the next from \citet{HM}, and IndigoBM from \citet{Indigo}.}
\label{tab:inductive_entity_datasets}
\end{table*}

\begin{table*}[ht]
\centering
\small
\begin{tabular}{l ccc | cccc | cccc}
\toprule
\multirow{2}{*}{\textbf{Dataset}} 
& \multicolumn{3}{c}{\textbf{Train Graph}} 
& \multicolumn{4}{c}{\textbf{Validation Graph}} 
& \multicolumn{4}{c}{\textbf{Test Graph}} \\
\cmidrule(lr){2-4} \cmidrule(lr){5-8} \cmidrule(lr){9-12}
& \textbf{$|\mathcal{E}|$} & \textbf{$|\mathcal{R}|$} & \textbf{$|\mathcal{F}|$} 
& \textbf{$|\mathcal{E}|$} & \textbf{$|\mathcal{R}|$} & \textbf{$|\mathcal{F}|$} & \textbf{Valid} 
& \textbf{$|\mathcal{E}|$} & \textbf{$|\mathcal{R}|$} & \textbf{$|\mathcal{F}|$} & \textbf{Test} \\
\midrule
FB-25           & 5190  & 163 & 91571  & 4097  & 216 & 17147 & 5716  & 4097  & 216 & 17147 & 5716 \\
FB-50           & 5190  & 153 & 85375  & 4445  & 205 & 11636 & 3879  & 4445  & 205 & 11636 & 3879 \\
FB-75           & 4659  & 134 & 62809  & 2792  & 186 & 9316  & 3106  & 2792  & 186 & 9316  & 3106 \\
FB-100          & 4659  & 134 & 62809  & 2624  & 77  & 6987  & 2329  & 2624  & 77  & 6987  & 2329 \\
WK-25           & 12659 & 47  & 41873  & 3228  & 74  & 3391  & 1130  & 3228  & 74  & 3391  & 1131 \\
WK-50           & 12022 & 72  & 82481  & 9328  & 93  & 9672  & 3224  & 9328  & 93  & 9672  & 3225 \\
WK-75           & 6853  & 52  & 28741  & 2722  & 65  & 3430  & 1143  & 2722  & 65  & 3430  & 1144 \\
WK-100          & 9784  & 67  & 49875  & 12136 & 37  & 13487 & 4496  & 12136 & 37  & 13487 & 4496 \\
NL-0            & 1814  & 134 & 7796   & 2026  & 112 & 2287  & 763   & 2026  & 112 & 2287  & 763 \\
NL-25           & 4396  & 106 & 17578  & 2146  & 120 & 2230  & 743   & 2146  & 120 & 2230  & 744 \\
NL-50           & 4396  & 106 & 17578  & 2335  & 119 & 2576  & 859   & 2335  & 119 & 2576  & 859 \\
NL-75           & 2607  & 96  & 11058  & 1578  & 116 & 1818  & 606   & 1578  & 116 & 1818  & 607 \\
NL-100          & 1258  & 55  & 7832   & 1709  & 53  & 2378  & 793   & 1709  & 53  & 2378  & 793 \\
\midrule
Metafam         & 1316  & 28  & 13821  & 1316  & 28  & 13821 & 590   & 656   & 28  & 7257  & 184 \\
FBNELL          & 4636  & 100 & 10275  & 4636  & 100 & 10275 & 1055  & 4752  & 183 & 10685 & 597 \\
MT1 tax    & 10000 & 10  & 17178  & 10000 & 10  & 17178 & 1908  & 10000 & 9   & 16526 & 1834 \\
MT1 health  & 10000 & 7   & 14371  & 10000 & 7   & 14371 & 1596  & 10000 & 7   & 14110 & 1566 \\
MT2 org     & 10000 & 10  & 23233  & 10000 & 10  & 23233 & 2581  & 10000 & 11  & 21976 & 2441 \\
MT2 sci    & 10000 & 16  & 16471  & 10000 & 16  & 16471 & 1830  & 10000 & 16  & 14852 & 1650 \\
MT3 art    & 10000 & 45  & 27262  & 10000 & 45  & 27262 & 3026  & 10000 & 45  & 28023 & 3113 \\
MT3 infra  & 10000 & 24  & 21990  & 10000 & 24  & 21990 & 2443  & 10000 & 27  & 21646 & 2405 \\
MT4 sci    & 10000 & 42  & 12576  & 10000 & 42  & 12576 & 1397  & 10000 & 42  & 12516 & 1388 \\
MT4 health  & 10000 & 21  & 15539  & 10000 & 21  & 15539 & 1725  & 10000 & 20  & 15337 & 1703 \\
\bottomrule
\end{tabular}
\caption{Dataset-wise statistics for the 23 datasets used under the inductive entity and relation ($e,r$) generalization regime. \textbf{$|\mathcal{R}|$} and \textbf{$|\mathcal{E}|$} indicate vocabulary sizes of relations and entities present in each split, respectively. \textbf{Train}, \textbf{Validation}, and \textbf{Test Graphs} include the triples present in each split. The \textbf{Test} and \textbf{Valid} columns denote the number of link prediction queries evaluated in each corresponding graph. The first half of the datasets are from \citet{ingram} and the second half from \citet{metafam}.}
\label{tab:inductive_entity_relation_datasets}
\end{table*}

\section{Leakage}
\label{appendix:Leakage}

During our evaluation of $\ultra$, we discovered a notable source of information leakage arising from the overlap between test sets and the pretraining corpus corroborating with findings from \citet{leakage}. Specifically, we found that 30 out of the 54 datasets used in the evaluation of $\ultra$ contain at least one triple from their test graph that is already present in the pretraining corpus. This overlap stems from the fact that many of the benchmark datasets in the 57-dataset corpus are derived from widely-used and well-known knowledge graphs such as WordNet, Freebase, Wikidata, and NELL, some of which also contribute to the pretraining data. We categorize this leakage into two distinct types:
\begin{enumerate}
    \item \textbf{Test Graph Leakage (Indirect):} This occurs when one or more triples from $\Gtest$ are already present in the pretraining corpus. Although the specific query triple may not be included, the model can potentially exploit these known graph facts to better predict test queries, thus providing an unfair advantage. This represents an indirect form of information leakage.

    \item \textbf{Query Leakage (Direct):} This refers to a more explicit form of leakage where the exact test query triples themselves are present in the pretraining corpus. In such cases, the model may have already been trained to predict the ground truth, effectively reducing the evaluation to a memorization check rather than a generalization task.
\end{enumerate}

\begin{figure*}[h]
    \centering
    \includegraphics[width=\linewidth]{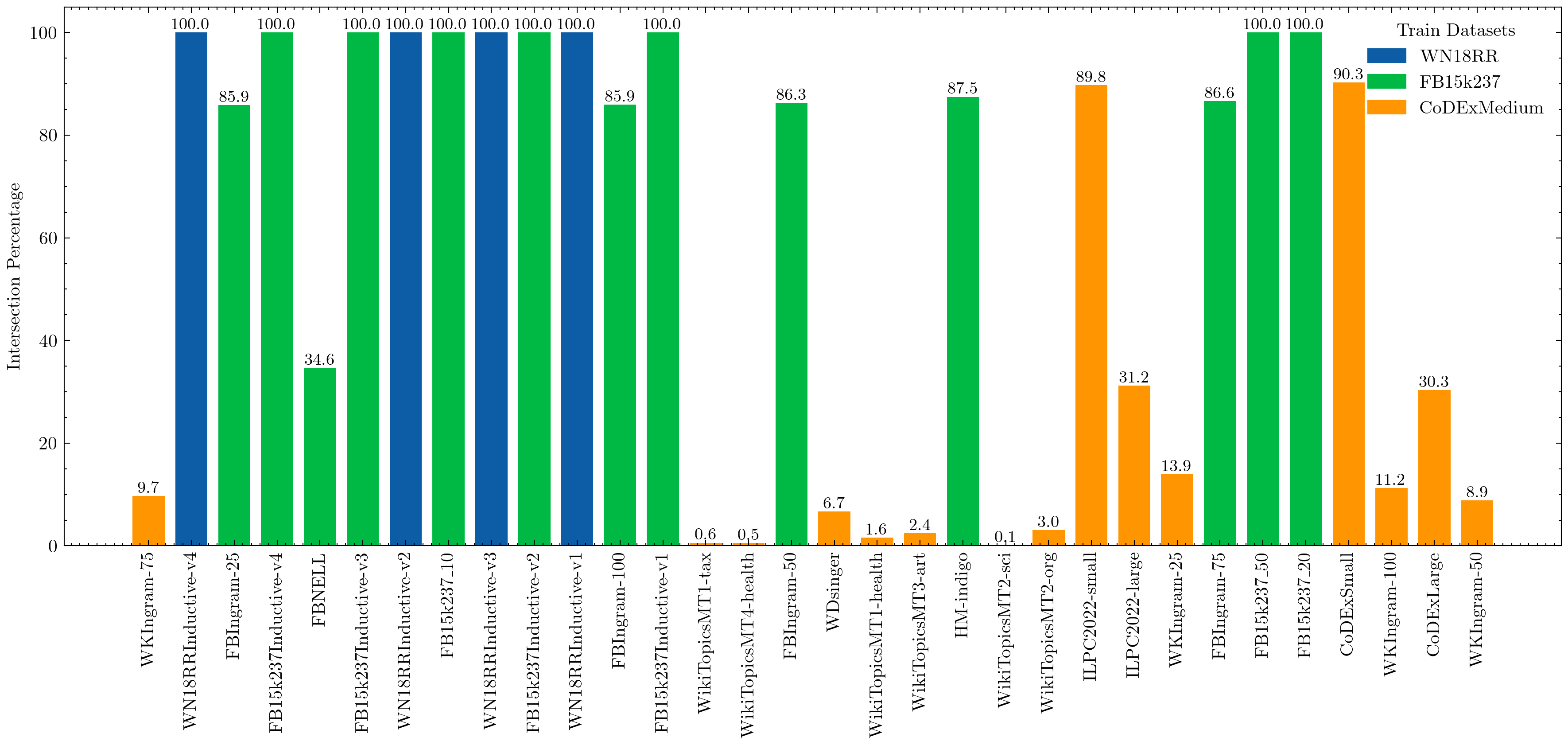}
    \caption{\textbf{Test graph Leak.} Percentage of test graph triples found in the pretraining corpus, indicating indirect leakage across datasets. Colors represent the corresponding training datasets in which leakage was found.}
    \label{fig:inf-leak}
\end{figure*}

\begin{figure*}[h]
    \centering
    \includegraphics[width=\linewidth]{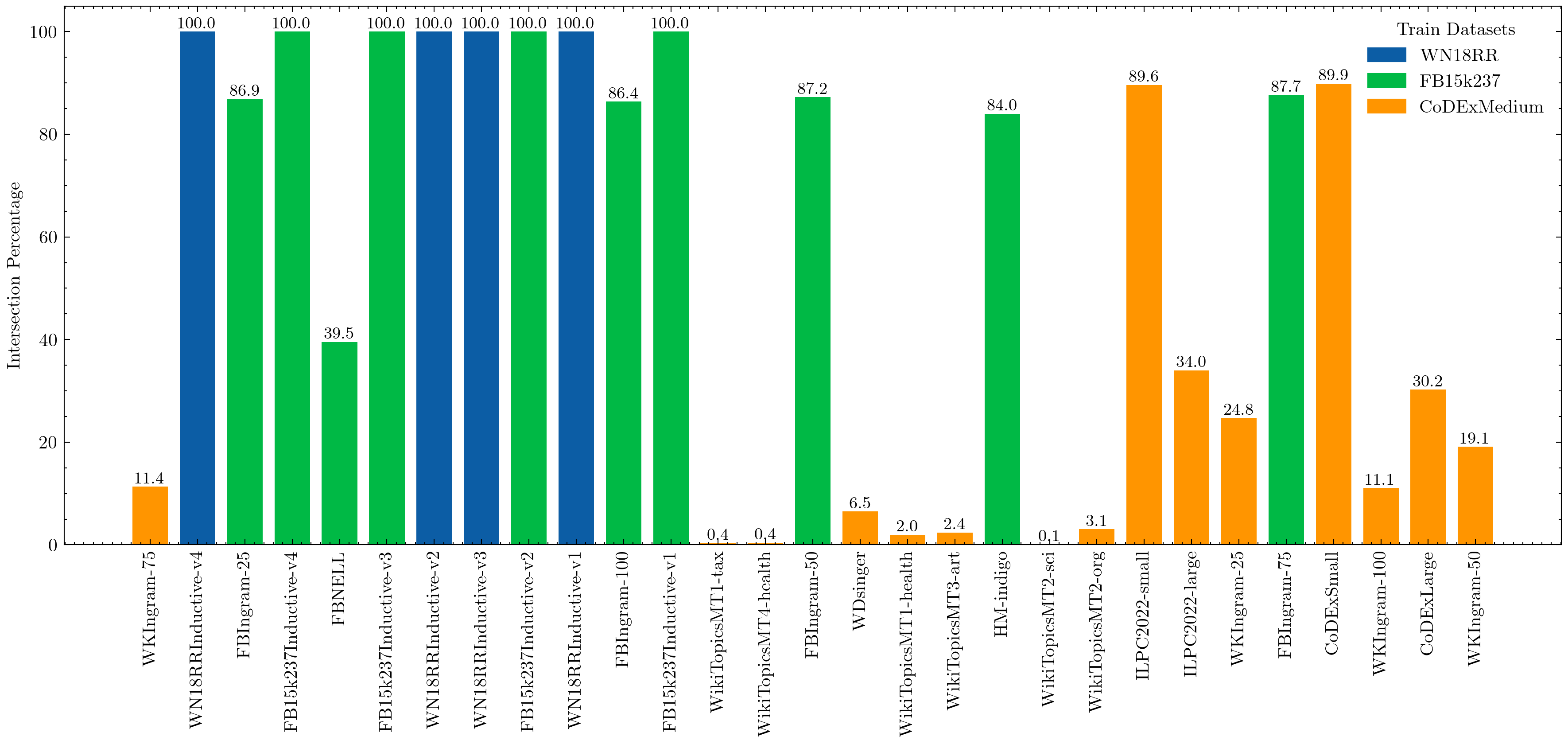}
    \caption{\textbf{Query triple Leak.} Percentage of query triples found in the pretraining corpus, indicating direct leakage across datasets. Colors represent the corresponding training datasets in which leakage was found.}
    \label{fig:test-leak}
\end{figure*}

We conduct a detailed analysis of both types of leakage and quantify their extent across each dataset. The results are presented in \Cref{fig:inf-leak} and \Cref{fig:test-leak}, which respectively depict the proportion of datasets affected by indirect (test graph) and direct (query triple) leakage. This analysis highlights the importance of careful dataset curation and motivates the need for evaluating models in settings where such overlaps are explicitly mitigated.

\section{Implementation Details}
\label{appendix:Implementation}

\textbf{Codebase.} Our implementation is based on the official $\ultra$ codebase,\footnote{https://github.com/DeepGraphLearning/ULTRA} which we extended to integrate the proposed dual-module architecture, including the LLM-based relation enrichment, $\rgtwo$ construction, and the structural-textual fusion module. The full source code for $\ours$ and a pre-trained model checkpoint are available in the supplementary material.

\noindent \textbf{Hyperparameters.} For the structure processing module (NBFNet operating on $\rgone$) and the final entity-level NBFNet, we largely retain the hyperparameter configurations reported by $\ultra$ (see \Cref{tab:pretrain_hpo} for a summary). Hyperparameters specific to $\ours$'s novel components were determined as follows: the $\rgtwo$ construction threshold was set to 0.8 based on ablation studies and \citet{threshold}; the MLP for fusion ($F$) consists of one hidden layer of size 128 with ReLU activation, chosen based on preliminary experiments on a validation subset of the pre-training data. The choice of LLM (gpt-4o) and text embedding model (Jina-embeddings-v3) was also informed by our ablations (\Cref{tab:ablations}). We only explore late fusion techniques as we want to reserve the option to switch off the text processing module.

\noindent \textbf{Training \& Complexity.} $\ours$ has a total of approximately 227k trainable parameters. The pre-training phase, conducted on the combined pre-training dataset of 500K samples, required approximately 9 hours on a single NVIDIA A100 GPU for 10 epochs. At inference time, the generation of LLM-enriched representations for the relation vocabulary $\mathcal{R}$ introduces an additional computational step with complexity proportional to $O(|\mathcal{R}| \times\ L_{\text{LLM}})$, where $L_{\text{LLM}}$ is the cost of LLM processing per relation. The subsequent GNN inference on $\rgtwo$ and $\rgone$ remains efficient.

We also utilized AI assistants during the development cycle for tasks such as code debugging and suggesting alternative phrasing for manuscript clarity, ensuring the core contributions and final methodology remained our own.

\begin{table*}[!htp]
\centering
\begin{tabular}{lccc}\toprule
Component & Hyperparameter & Value \\\midrule
\multirow{4}{*}{$\text{GNN}_\textsc{str}$} &\# layers &6 \\
&hidden dim &64 \\
&message &DistMult \\
&aggeregation &sum \\ \midrule
\multirow{4}{*}{$\text{GNN}_\textsc{text}$} &\# layers &6 \\
&hidden dim &64 \\
&message &DistMult \\
&aggeregation &sum \\ \midrule
\multirow{5}{*}{$\text{GNN}_\textsc{ent}$} &\# layers &6 \\
&hidden dim &64 \\
&message &DistMult \\
&aggregation &sum \\
&$g(\cdot)$ &2-layer MLP \\ \midrule
\multirow{7}{*}{Learning} &optimizer &AdamW \\
&learning rate &0.0005 \\
&training steps &800,000 \\
&adv temperature &1 \\
&\# negatives &512 \\
&batch size &64 \\
&Training graph mixture &FB15k237, WN18RR, CoDExMedium \\
\bottomrule
\end{tabular}
\caption{\textbf{$\ours$ hyperparameters for pre-training.} $\text{GNN}_\textsc{str}$ corresponds to the NBFNet that operates on $\rgone$, $\text{GNN}_\textsc{sem}$ to the NBFNet that operates on $\rgtwo$ and $\text{GNN}_\textsc{ent}$ to the NBFNet that operates on entity level.}
\label{tab:pretrain_hpo}
\end{table*}

\section{Performance variance in \texorpdfstring{$\ours$}{OURS}}
\label{appendix:performance_variance}

The observed performance variations of our approach across different KGs may be attributed to their intrinsic structural and textual properties. KGs such as Hetionet and YAGO310 seem to be carefully curated with standardized relational schemas, where each relation possesses precise semantics and consistent usage. In such highly regularized environments, the introduction of textual semantic similarity mechanisms, if not carefully calibrated, could potentially disrupt this precision, leading to an over-smoothing effect and a degradation in reasoning performance by failing to preserve the subtle semantic differences that distinguish various relations.

Conversely, datasets like ConceptNet, which incorporates commonsense knowledge, and Metafam, exhibit relations with comparatively looser standardization and greater linguistic variability. In these contexts, our textual semantic similarity approach appears to effectively identify important semantic parallels, thereby bridging linguistic variations and compensating for less formal relational definitions.

The usefulness of textual semantic similarity is also dependent upon the threshold used. A high threshold predominantly captures high-certainty similarities (e.g., exact synonymy, close paraphrases), while lower thresholds may encompass broader contextual or domain-specific similarities. This leads to a hypothesis: highly standardized KGs might benefit from a stringent threshold to preserve relational specificity, whereas KGs with more flexible relational definitions could achieve optimal performance with a more lenient threshold that accommodates linguistic diversity. Based on preliminary experiments, we also observed a correlation between the sparsity of the KG and the value of textual semantics. While this provides a high-level rationale, we acknowledge that other factors, such as the distribution of relation types and graph topology, likely influence the observed outcomes.

So, from these preliminary analyses, we hypothesize $\ours$ is especially better for datasets where there is a significant deviance between the structure and the textual semantics, or if the relation text is rich, or if the KG is sparse.

\noindent \textbf{Sensitivity Analysis.} We observed that the model's performance is sensitive to the cosine similarity threshold used for edge creation in $\rgtwo$. Consequently, we performed a grid search over the values \{0.6, 0.7, 0.79, 0.8\} as part of our hyperparameter tuning process. Based on this search and prior works \cite{threshold}, we selected a final threshold of 0.8 and further validated this choice by conducting a qualitative analysis of the generated edges on a subset of the datasets. Investigating dynamic thresholding techniques for different datasets remains a promising direction for future work.

\section{Harder setting}
\label{appendix:harder}
We extend our evaluation to a more challenging setting, wherein the relation vocabulary of queries is entirely disjoint from the relation vocabulary present in the test graph. Formally, given $\Gtest$ and queries $Q = \Ftruth \setminus \Fobs$, we define a new evaluation setting:
\begin{equation*}
\mathcal{R}_{Q} \cap \mathcal{R}_{\Gtest} = \emptyset
\end{equation*}
where $\mathcal{R}_{Q}$ denotes the set of relations used in queries, and $\mathcal{R}_{\Gtest}$ denotes the set of relations present in the test graph. This scenario is motivated by the practical case of temporally evolving knowledge graphs, wherein new relations frequently emerge over time and must be incorporated dynamically into existing inference frameworks \citep{takeaway3}. To construct datasets compliant with this harder setting, we perform the following procedure:
\begin{enumerate}
    \item Start with the original dataset split, combining the test graph and test triples to create a unified set of facts.
    \item Randomly select and mask out a subset of relations from the relation vocabulary of the test graph, based on a predefined split ratio.
    \item Remove all triples involving these masked relations from the combined set, effectively filtering out these relations from the test graph.
    \item From the filtered combined set, we define a subset to serve as our new test graph.
    \item From the masked-out set of triples (associated with the masked relations), we select a representative subset as new test triples, ensuring the ratio of test graph to test triples matches the original dataset split and that entities in the test triples appear in the test graph.
\end{enumerate}

In this more stringent setting, purely structural approaches such as $\ultra$ encounter significant difficulties. Specifically, when multiple distinct relations in the test triples share the same head or tail entity, $\ultra$ fails to differentiate among these relations. This limitation arises because $\ultra$ relies solely on structural identifiers (IDs), assigning identical temporary IDs to unseen relations at inference time due to their absence from the test graph. 

In contrast, $\ours$ effectively addresses this limitation. By incorporating semantic embeddings derived from textual descriptions, $\ours$ distinguishes between novel, previously unseen relations, leveraging the rich semantic signals inherent in relation texts. Consequently, $\ours$ achieves significantly better prediction performance compared to purely structural methods as shown in \Cref{tab:takeaway3}.

\begin{figure*}[h]
    \centering
    \includegraphics[width=\linewidth]{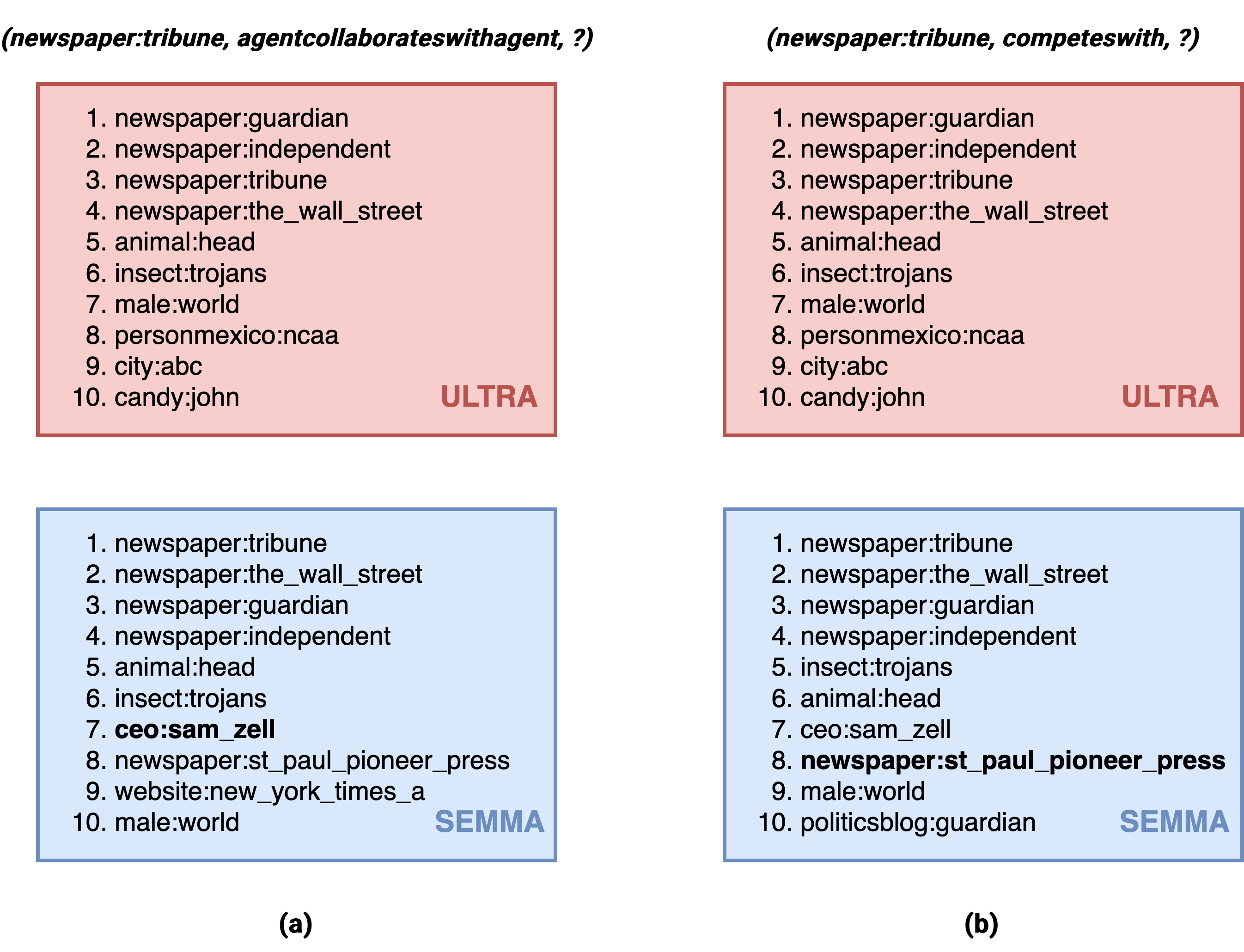}
    \caption{Comparison of $\ultra$ and $\ours$ in the challenging setting where the query triples relation vocabulary is disjoint from the test graph relation vocabulary. $\ultra$ fails to differentiate between distinct relations (\texttt{agentcollaborateswithagent} vs. \texttt{competeswith}), producing identical incorrect predictions. In contrast, $\ours$ distinguishes between the two relations and correctly predicts the ground truth within its top-10 predictions.}
    \label{fig:t3-example}
\end{figure*}

Consider the example shown in \Cref{fig:t3-example}, where $\ultra$ is unable to distinguish between \texttt{agentcollaborateswithagent} and \texttt{competeswith}, resulting in identical and incorrect top-10 predictions for both relations. In contrast, $\ours$ successfully differentiates these two relations, as evident from the clearly ordered top-10 predictions that include the correct ground truth. This clearly illustrates $\ours$'s ability to leverage textual understanding to overcome structural ambiguities.

\section{Full results}
\label{appendix:fullresults}
We provide the regime-wise averages of the dataset for our ablation studies in \Cref{tab:LLM-full}. The complete results per data set reporting MRR and Hits@10 of the zero-shot inference of the pre-trained $\ultra$ and $\ours$ model are presented in \Cref{tab:tabl1-half1} and \Cref{tab:table1-half2}.

\section{Discussion on Parallel works}
\label{appendix:parallel-works}

REEF \cite{yu2025relationawaregraphfoundationmodel} leverages ``relation tokens'' from textual descriptions to adaptively generate parameters for GNN components via hypernetworks, aiming for effective pre-training and transfer, while $\ours$ fuses distinct structural and textual relation graph representations for link prediction. SCR \cite{wang2025graphfoundationmodelstraining} trains on KGs and reformulates general graph tasks into an inductive KG reasoning format to enable transfer, proposing Semantic Conditional Message Passing. SCR focuses on transferring KG reasoning to general graph tasks by reformatting them, whereas $\ours$ focuses on inductive link prediction \textit{within} KGs using relation semantics.

KumoRFM \cite{kumo2025kumorfm} is a pre-trained model for in-context learning on general relational databases, designed for zero-shot predictions across diverse enterprise tasks without specific training, using a Predictive Query Language. Insights like the helpfulness of textual semantics (\Cref{sec:rq2}) can potentially help improve such relational foundation models by leveraging the textual information in the database key and table names.

\begin{table*}[!t]
\small
\centering
\begin{tabular}{c l c c c c c c || c c}
\toprule
 & & \multicolumn{2}{c}{\textbf{Inductive $e, r$}} & \multicolumn{2}{c}{\textbf{Inductive $e$}} & \multicolumn{2}{c}{\textbf{Transductive}} & \multicolumn{2}{c}{\textbf{Total Avg}}\\

& & \multicolumn{2}{c}{(23 graphs)} & \multicolumn{2}{c}{(18 graphs)} & \multicolumn{2}{c}{(13 graphs)} & \multicolumn{2}{c}{(57 graphs)}\\
\cmidrule{3-10}
&  & \textbf{MRR} & \textbf{H@10} & \textbf{MRR} & \textbf{H@10} & \textbf{MRR} & \textbf{H@10} & \textbf{MRR} & \textbf{H@10} \\
\midrule
\multirow{3}{*}{\rotatebox{90}{\textbf{LLMs}}} & gpt-4o-2024-11-20                        & \bm{$0.355$} & $0.515$ & $0.448$ & $0.585$ & \bm{$0.322$} & \bm{$0.473$} & \bm{$0.377$} & $0.529$ \\
& deepseek-chat-v3-0324      & \bm{$0.355$} & \bm{$0.524$} & \bm{$0.45$} & \bm{$0.586$} & $0.307$ & $0.461$ & $0.375$ & \bm{$0.53$}\\
& qwen3-32b      & $0.343$ & $0.513$ & $0.44$ & $0.57$ & $0.313$ & $0.461$ & $0.368$ & $0.519$\\
\midrule
\multirow{2}{*}{\rotatebox{90}{\textbf{LM}}} & jina-embeddings-v3                        & \bm{$0.355$} & $0.515$ & \bm{$0.448$} & \bm{$0.585$} & \bm{$0.322$} & \bm{$0.473$} & \bm{$0.377$} & \bm{$0.529$} \\
& Sentence-BERT      & $0.339$ & \bm{$0.521$} & \bm{$0.448$} & $0.584$ & $0.308$ & $0.461$ & $0.368$ & $0.528$\\
\midrule
\multirow{2}{*}{\rotatebox{90}{\textbf{$F$}}} & MLP                        & \bm{$0.355$} & $0.515$ & \bm{$0.448$} & \bm{$0.585$} & \bm{$0.322$} & \bm{$0.473$} & \bm{$0.377$} & \bm{$0.529$} \\
& Attention      & $0.343$ & \bm{$0.518$} & $0.441$ & $0.577$ & $0.314$ & $0.467$ & $0.369$ & $0.525$\\
\midrule
\multirow{5}{*}{\rotatebox{90}{\textbf{Text}}} & $\textsc{rel\_name}$ & $0.338$ & $0.517$ & \bm{$0.449$} & \bm{$0.588$} & $0.314$ & $0.465$ & $0.369$ & $0.528$\\
& $\textsc{llm\_rel\_name}$ & \bm{$0.358$} & \bm{$0.528$} & $0.447$ & $0.585$ & $0.308$ & $0.456$ & $0.375$ & \bm{$0.529$}\\
& $\textsc{llm\_rel\_desc}$ & \bm{$0.358$} & $0.525$ & $0.436$ & $0.566$ & $0.312$ & $0.4625$ & $0.373$ & $0.524$\\
& \cs                      & $0.355$ & $0.515$ & $0.448$ & $0.585$ & \bm{$0.322$} & \bm{$0.473$} & \bm{$0.377$} & \bm{$0.529$} \\
& \cavg                      & $0.348$ & $0.507$ & $0.45$ & $0.584$ & $0.316$ & $0.466$ & $0.374$ & $0.523$\\
\midrule
\multirow{2}{*}{\smash{\rotatebox[origin=c]{90}{\textbf{$\rgtwo$}}}} & Threshold (0.8)                        & \bm{$0.355$} & \bm{$0.515$} & $0.448$ & $0.585$ & \bm{$0.322$} & \bm{$0.473$} & \bm{$0.377$} & \bm{$0.529$} \\
& Top-x\% (20\%)  & $0.341$ & $0.509$ & \bm{$0.453$} & \bm{$0.591$} & $0.31$ & $0.462$ & $0.371$ & $0.525$\\
\bottomrule
\end{tabular}
\caption{\textbf{Ablation Studies.} Evaluating the impact of different design choices in \textsc{$\ours$}. We report MRR and Hits@10 across the three evaluation regimes. The table compares (i) different LLMs, (ii) language encoders for deriving relation embeddings, (iii) fusion mechanisms (MLP vs. Attention), (iv) variations in relation textual input, and (v) different strategies for constructing the textual relation graph ($\rgtwo$).}
\label{tab:LLM-full}
\end{table*}

\begin{table*}[!ht]
\centering
\begin{tabular}{lcccc}
\toprule
\multirow{2}{*}{\textbf{Dataset}} & \multicolumn{2}{c}{\textbf{$\ultra$}} & \multicolumn{2}{c}{\textbf{$\ours$}} \\ \cmidrule(l){2-3} \cmidrule(l){4-5} 
 & \textbf{MRR} & \textbf{H@10} & \textbf{MRR} & \textbf{H@10} \\
\midrule
\multicolumn{5}{c}{\textit{Pre-training datasets}} \\
\midrule
FB15k237 & 0.372 & 0.569 & \textbf{0.377} & \textbf{0.574} \\
WN18RR & 0.499 & 0.622 & \textbf{0.548} & \textbf{0.656} \\
CoDExMedium & \textbf{0.377} & \textbf{0.531} & 0.376 & 0.529 \\
\midrule
\multicolumn{5}{c}{\textit{Transductive datasets}} \\
\midrule
CoDExSmall & 0.470 & \textbf{0.674} & \textbf{0.479} & 0.672 \\
CoDExLarge & 0.339 & 0.470 & \textbf{0.349} & \textbf{0.478} \\
NELL995 & 0.402 & 0.534 & \textbf{0.442} & \textbf{0.577} \\
DBpedia100k & 0.387 & 0.561 & \textbf{0.408} & \textbf{0.576} \\
ConceptNet100k & 0.109 & 0.207 & \textbf{0.162} & \textbf{0.311} \\
NELL23k & 0.234 & 0.390 & \textbf{0.242} & \textbf{0.413} \\
YAGO310 & \textbf{0.467} & \textbf{0.634} & 0.394 & 0.567 \\
Hetionet & \textbf{0.289} & \textbf{0.412} & 0.249 & 0.361 \\
WDsinger & 0.365 & 0.480 & \textbf{0.386} & \textbf{0.496} \\
AristoV4 & 0.223 & 0.322 & \textbf{0.232} & \textbf{0.346} \\
FB15k237\_10 & 0.243 & 0.391 & \textbf{0.244} & \textbf{0.393} \\
FB15k237\_20 & 0.269 & \textbf{0.434} & \textbf{0.269} & 0.430 \\
FB15k237\_50 & 0.324 & \textbf{0.525} & \textbf{0.324} & 0.525 \\
\midrule
\multicolumn{5}{c}{\textit{Inductive (e) datasets}} \\
\midrule
FB15k237Inductive:v1 & 0.481 & 0.647 & \textbf{0.486} & \textbf{0.655} \\
FB15k237Inductive:v2 & 0.493 & 0.682 & \textbf{0.503} & \textbf{0.691} \\
FB15k237Inductive:v3 & 0.480 & 0.641 & \textbf{0.494} & \textbf{0.651} \\
FB15k237Inductive:v4 & 0.476 & 0.664 & \textbf{0.492} & \textbf{0.677} \\
WN18RRInductive:v1 & 0.615 & 0.767 & \textbf{0.724} & \textbf{0.816} \\
WN18RRInductive:v2 & 0.658 & 0.766 & \textbf{0.705} & \textbf{0.803} \\
WN18RRInductive:v3 & 0.379 & 0.491 & \textbf{0.442} & \textbf{0.577} \\
WN18RRInductive:v4 & 0.598 & 0.712 & \textbf{0.664} & \textbf{0.741} \\
NELLInductive:v1 & 0.732 & 0.863 & \textbf{0.798} & \textbf{0.935} \\
NELLInductive:v2 & 0.501 & 0.698 & \textbf{0.543} & \textbf{0.730} \\
NELLInductive:v3 & 0.509 & 0.680 & \textbf{0.530} & \textbf{0.720} \\
NELLInductive:v4 & 0.476 & 0.703 & \textbf{0.496} & \textbf{0.729} \\
ILPC2022:small & \textbf{0.299} & 0.447 & 0.298 & \textbf{0.449} \\
ILPC2022:large & 0.297 & 0.422 & \textbf{0.307} & \textbf{0.429} \\
HM:1k & \textbf{0.063} & 0.105 & 0.062 & \textbf{0.109} \\
HM:3k & 0.052 & 0.098 & \textbf{0.056} & \textbf{0.102} \\
HM:5k & 0.047 & 0.088 & \textbf{0.055} & \textbf{0.102} \\
HM:indigo & \textbf{0.439} & \textbf{0.651} & 0.435 & 0.645 \\
\bottomrule
\end{tabular}
\caption{Comparison of $\ultra$ and $\ours$ across transductive and partially inductive regime. The bold indicates the highest value of that metric for a specific dataset.}
\label{tab:tabl1-half1}
\end{table*}

\begin{table*}[!ht]
\centering
\begin{tabular}{lcccc}
\toprule
\multirow{2}{*}{\textbf{Dataset}} & \multicolumn{2}{c}{\textbf{$\ultra$}} & \multicolumn{2}{c}{\textbf{$\ours$}} \\ \cmidrule(l){2-3} \cmidrule(l){4-5} 
 & \textbf{MRR} & \textbf{H@10} & \textbf{MRR} & \textbf{H@10} \\
\midrule
\multicolumn{5}{c}{\textit{Inductive (e, r) datasets}} \\
\midrule
FBIngram:25 & 0.383 & 0.636 & \textbf{0.400} & \textbf{0.642} \\
FBIngram:50 & 0.332 & 0.536 & \textbf{0.344} & \textbf{0.546} \\
FBIngram:75 & 0.395 & 0.598 & \textbf{0.404} & \textbf{0.600} \\
FBIngram:100 & 0.436 & 0.634 & \textbf{0.445} & \textbf{0.635} \\
WKIngram:25 & 0.288 & 0.481 & \textbf{0.303} & \textbf{0.509} \\
WKIngram:50 & 0.152 & 0.304 & \textbf{0.174} & \textbf{0.318} \\
WKIngram:75 & 0.372 & \textbf{0.534} & \textbf{0.387} & 0.525 \\
WKIngram:100 & 0.178 & 0.295 & \textbf{0.179} & \textbf{0.301} \\
NLIngram:0 & 0.327 & 0.491 & \textbf{0.367} & \textbf{0.567} \\
NLIngram:25 & 0.381 & 0.534 & \textbf{0.387} & \textbf{0.548} \\
NLIngram:50 & 0.365 & 0.531 & \textbf{0.409} & \textbf{0.574} \\
NLIngram:75 & 0.333 & 0.494 & \textbf{0.353} & \textbf{0.544} \\
NLIngram:100 & 0.444 & 0.631 & \textbf{0.465} & \textbf{0.676} \\
WikiTopicsMT1:tax & \textbf{0.232} & \textbf{0.302} & 0.229 & 0.302 \\
WikiTopicsMT1:health & 0.308 & 0.419 & \textbf{0.336} & \textbf{0.435} \\
WikiTopicsMT2:org & 0.086 & 0.146 & \textbf{0.096} & \textbf{0.158} \\
WikiTopicsMT2:sci & \textbf{0.270} & \textbf{0.427} & 0.258 & 0.388 \\
WikiTopicsMT3:art & 0.270 & \textbf{0.418} & \textbf{0.278} & 0.416 \\
WikiTopicsMT3:infra & 0.637 & 0.777 & \textbf{0.650} & \textbf{0.784} \\
WikiTopicsMT4:sci & \textbf{0.288} & 0.449 & 0.287 & \textbf{0.454} \\
WikiTopicsMT4:health & 0.580 & 0.735 & \textbf{0.615} & \textbf{0.739} \\
Metafam & 0.155 & \textbf{0.565} & \textbf{0.258} & 0.530 \\
FBNELL & 0.474 & 0.638 & \textbf{0.482} & \textbf{0.655} \\
\bottomrule
\end{tabular}
\caption{Comparison of $\ultra$ and $\ours$ across fully inductive regime. The bold indicates the highest value of that metric for a specific dataset.}
\label{tab:table1-half2}
\end{table*}

\end{document}